\documentclass[10pt,journal,compsoc]{IEEEtran}
\usepackage{amsmath,amsfonts}
\usepackage{algorithmic}
\usepackage{algorithm}
\usepackage{array}
\usepackage{xspace}
\usepackage{textcomp}
\usepackage{stfloats}
\usepackage{url}
\usepackage{verbatim}
\usepackage{graphicx}
\usepackage{cite}
\usepackage{hyperref}
\usepackage{booktabs}
\usepackage{multirow}
\usepackage{amssymb}
\usepackage{subfigure}
\usepackage{makecell}
\usepackage{soul}
\usepackage{xcolor}
\usepackage{xcolor,colortbl}
\usepackage{threeparttable}
\usepackage{tikz}
\usepackage[framemethod=TikZ]{mdframed}

\newcommand{\tool}[0]{\textsc{AI-Compass}\xspace}

\newcommand{\pie}[1]{%
\begin{tikzpicture}
 \draw (0ex,0ex) circle (1ex);
 \fill (0ex,-1ex) arc (-90:(#1-90):1ex) -- (0ex,-1ex) -- cycle;
\end{tikzpicture}%
}

\def\eg{\emph{e.g.,}\xspace}

\def\ie{\emph{i.e.,}\xspace}
\def\etal{\emph{et al.}\xspace}

\begin{document}

\title{AI-Compass: A Comprehensive and Effective Multi-module Testing Tool for AI Systems}

\author{\IEEEauthorblockN{Zhiyu Zhu\IEEEauthorrefmark{1},
Zhibo Jin\IEEEauthorrefmark{1},
Hongsheng Hu,
Minhui Xue,
Ruoxi Sun, \\
Seyit Camtepe, 
Praveen Gauravaram,
and
Huaming Chen\IEEEauthorrefmark{2}
}
\thanks{Zhiyu Zhu and Zhibo Jin are with the School of Electrical and Computer Engineering, The University of Sydney and CSIRO’s Data61 (e-mail: \{zzhu2018, zjin0915\}@uni.sydney.edu.au}
\thanks{Huaming Chen is with the School of Electrical and Computer Engineering, The University of Sydney (e-mail: \{huaming.chen, tj.lim\}@sydney.edu.au).}
\thanks{Hongsheng Hu, Minhui Xue, Ruoxi Sun, and Seyit Camtepe are with CSIRO's Data61 and Cybersecurity CRC (e-mail: \{hongsheng.hu, jason.xue, ruoxi.sun, seyit.camtepe\}@data61.csiro.au.}
\thanks{Praveen Gauravaram is with Cybersecurity CRC.}
\thanks{\IEEEauthorrefmark{1}Equal contribution.}
\thanks{\IEEEauthorrefmark{2}Corresponding author.}
}




\IEEEtitleabstractindextext{%
\begin{abstract}
AI systems, in particular with deep learning techniques, have demonstrated superior performance for various real-world applications. Given the need for tailored optimization in specific scenarios, as well as the concerns related to the exploits of subsurface vulnerabilities, a more comprehensive and in-depth testing AI system becomes a pivotal topic. We have seen the emergence of testing tools in real-world applications that aim to expand testing capabilities. However, they often concentrate on ad-hoc tasks, rendering them unsuitable for simultaneously testing multiple aspects or components. 
Furthermore, trustworthiness issues arising from adversarial attacks and the challenge of interpreting deep learning models pose new challenges for developing more comprehensive and in-depth AI system testing tools. 
In this study, we design and implement a testing tool, \tool, to comprehensively and effectively evaluate AI systems. The tool extensively assesses multiple measurements towards adversarial robustness, model interpretability, and performs neuron analysis. The feasibility of the proposed testing tool is thoroughly validated across various modalities, including image classification, object detection, and text classification. Extensive experiments demonstrate that \tool is the state-of-the-art tool for a comprehensive assessment of the robustness and trustworthiness of AI systems. Our research sheds light on a general solution for AI systems testing landscape.

\end{abstract}

}

\maketitle

\section{Introduction}
In recent years, the remarkable improvement of deep learning models have revolutionized the landscape of various industry sectors and application domains, showcasing their unparalleled potential in solving complex problems and driving innovation~\cite{he2016deep,simonyan2014very,xiong2016achieving,chen2020software,devanbu2020deep,li2021deeppayload,mirabella2021deep,sedaghatbaf2021automated}. The dynamic interplay between data-driven insights and sophisticated model architectures has propelled deep learning to the forefront of modern technology, enabling groundbreaking advancements across a myriad of novel applications~\cite{xiao2021development,shrestha2019review,parvat2017survey}. From enhancing medical diagnostics through image analysis to enabling autonomous vehicles to navigate and make informed decisions, the transformative capabilities of deep learning models have left an indelible mark on society~\cite{dhar2023challenges,kavitha2023ant,lee2023end,alaba2023deep}. To contextualize this transformative power, consider the case of natural language processing where models like GPT-3 have demonstrated human-level proficiency in generating coherent and contextually relevant text, ushering in a new era of interactive and responsive AI systems~\cite{haluza2023artificial,maddigan2023chat2vis}. Such remarkable feats underscore the urgent need for a comprehensive assessment framework that can holistically evaluate the multifaceted dimensions of deep learning models, delving into the intricate interplay of vast datasets, intricate model architectures, and immense computational resources that underpin their unprecedented success~\cite{zohdinasab2023efficient,sekhon2019towards,gerasimou2020importance}.

The comprehensive assessment of deployed machine learning (ML) models, particularly deep learning models, is of paramount importance~\cite{wadawadagi2020sentiment}. Such assessments serve as a crucial precautionary measure to uncover potential pitfalls and unanticipated consequences that could arise from utilizing inadequately evaluated models~\cite{jin2023novel,alwadi2023framework}. Conducting a thorough performance and security evaluation ensures a nuanced understanding of the models' capabilities, limitations, and potential biases, empowering informed decision-making and responsible deployment~\cite{guo2023comprehensive}. Neglecting a comprehensive assessment when deploying deep learning models can lead to detrimental outcomes. Biased predictions, unreliable results, and unexpected behaviors may emerge, eroding user trust, triggering legal and ethical challenges, and compromising the models' real-world performance~\cite{gajera2023comprehensive}. Real-world examples vividly illustrate these perils. In healthcare, deploying a poorly assessed AI diagnostic system could endanger patients through misdiagnoses\cite{sawhney2023comparative}. For instance, malicious actors may mislead the ML tumor detection system into erroneously classifying benign tumors as malignant ones by introducing imperceptible perturbations to the original medical images. This has the potential to misguide the physician's judgment, subsequently leading to irreversible harm to the patient's health~\cite{finlayson2018adversarial}. While inadequately evaluated autonomous vehicles might make flawed decisions, resulting in accidents~\cite{nesti2022evaluating}. For example, attackers can launch attacks on autonomous driving systems by introducing imperceptible perturbations to traffic signs. By applying imperceptible perturbations to stop signs as perceived by human eyes, the ML system may misclassify them as yield signs. This could lead to severe traffic accidents, posing a threat to user safety~\cite{gu2017badnets}. The financial sector is also at risk, as untested deep learning algorithms in market predictions could yield severe economic repercussions~\cite{mehtab2022analysis}. These instances underscore the urgency of a robust assessment framework, which is essential for mitigating risks and ensuring the safe and effective deployment of deep learning models~\cite{qiu2022multimodal}.

To achieve a comprehensive assessment of deep models, it is imperative to delve into three pivotal dimensions: adversarial robustness, model explainability, and neuron analysis~\cite{zhang2022adversarial,dieber2022novel,sajjad2022neuron}. Adversarial robustness stands as a bulwark, ensuring consistent performance even amid uncertainties and adversarial scenarios, thus fortifying its real-world applicability~\cite{zhou2022understanding}. Meanwhile, model explainability serves as a beacon of transparency, demystifying the decision-making process and fostering trust, especially in contexts where accountability is paramount~\cite{dai2022comprehensive}. Simultaneously, the intricate realm of neuron analysis grants us a profound understanding of the model's inner workings, at the level of individual neurons, elucidating the pathways of feature extraction and representation learning~\cite{guan2022explaining}. The convergence of these facets not only empowers a comprehensive evaluation but also equips stakeholders with the insights needed to navigate the nuanced landscape of deep learning models, promoting informed deployment and harnessing their transformative potential across diverse domains~\cite{yang2022revisiting}.


In order to enhance the capability for testing Deep Learning Systems (DLS) in real-world applications, numeric testing tools have been developed.
Taking medical image analysis as an example, DLTK~\cite{pawlowski2017dltk}, as an open-source DL toolkit, provides a range of tools for testing and validating the quality of DLS, including model evaluation, model interpretation, and model visualization. DLTK provides a detailed diagnostic report for medical images, reducing the risk of misjudgment by explaining the model's behavior. DeepXplore~\cite{pei2017deepxplore} is an automated white-box testing framework for DLS that employs optimization techniques such as gradient ascent to detect potential failures in the system. As an effective approach for automated testing of deep neural network (DNN)-driven autonomous cars, DeepTest~\cite{tian2018deeptest} designs a test generation framework that combines mutation operators, metamorphic relations, and real-world driving scenarios to generate test cases with higher neuron coverage. 
However, even though the DLS testing tools are constantly being upgraded, resembling an arms race in multiple fields as described above, the inherent ad-hoc, task-oriented nature of existing tools persists as an unavoidable limitation, often making them unsuitable for fulfilling multi-task testing requirements.
For example, the testing objective of DeepXplore is monotonous, and it is exclusively applicable to white-box testing, which makes it unsuitable as a general-purpose testing tool in a black-box environment. Furthermore, DeepXplore does not provide an explanation for how a model's defects are detected. Both DLTK and DeepTest have limited testing capabilities in application scenarios unrelated to medical image analysis and DNN-driven autonomous driving, lacking sufficient tests of adversarial robustness or model interpretability. 
As far as we know, existing testing tools can only conduct individual module tests on a model's adversarial robustness, interpretability, or neuron analysis, rather than explaining the relationships among these three aspects. In order to enable multidimensional evaluation and selection of models, we are dedicated to integrating these modules for multi-task testing and constructing a comprehensive testing tool.
In addition, pruning has been proven to facilitate the interpretation of model decisions and reduce the occurrence of overfitting during adversarial sample training~\cite{weber2023less,jordao2021effect}. For the first time, we introduce an approach to neuron analysis with pruning techniques, thereby exploring potential connections among the modules.

In this paper, we propose \tool, a comprehensive and effective multi-module testing tool for DLS. 
Specifically, combined with the basic utility module including indicator evaluation and mutation operations~\cite{ma2018deepmutation,hu2019deepmutation++}, for the first time, we design modules for adversarial robustness, model interpretability and neuron analysis, to extensively evaluate the performance of DLS~\cite{jin2023poster}. 
Through a thorough validation involving 6 deep learning models across 3 datasets, we demonstrate that \tool is capable of testing image classification, object detection, and text classification tasks in DLS. 
Compared to existing DLS testing tools, \tool not only conducts fundamental DLS testing but also delivers precise evaluations of model robustness against adversarial attacks. Furthermore, it provides trustworthy model interpretability reports, including a quantified assessment of the tested model's interpretability, along with attributional result charts for illustration. 

The main contributions of this paper are as follows:
\begin{itemize}
    \item We present a comprehensive and effective framework, \tool, for automatically testing the quality of DLS. Specifically, combined with the basic utility module, for the first time, we design modules for adversarial robustness, model interpretability, and neuron analysis, making a significant step towards building robust and trustworthy DLS.
    \item Inspired by the pruning algorithm, we conduct an in-depth analysis of neural network redundancy. We comprehensively investigate the changes in adversarial robustness and model interpretability resulting from neuron pruning approach, thus providing valuable insights for model architecture optimization.
    \item We demonstrate that our \tool can be effectively applied for multi-modal scenarios. The testing results in image classification, text classification, and object detection tasks verify the high scalability of our \tool and solve the ad-hoc problem in existing testing tools.
    \item We have conducted extensive experiments and generated detailed test reports to demonstrate the superiority of our \tool in testing DLS.
    \item The code is released for future research and enhancements by scholars and industry professionals.
\end{itemize}

This study extends our previous conference paper~\cite{jin2023poster}. 
In Section~\ref{sec_related_work}, we provide an overview of related work on testing frameworks to afford readers a more comprehensive understanding of the field. Section~\ref{sec_preliminaries} introduces the preparatory background, furnishing foundational knowledge regarding adversarial attacks, model interpretability, and pruning algorithms. In Section~\ref{sec_structure_overview}, we expand the conceptual diagram of ML-compass~\cite{jin2023poster} to assist readers in gaining a clearer and more comprehensive grasp of the \tool architecture. 
Section~\ref{sec_methodology} delves into the methods and principles underpinning each module and sub-module to provide a more profound understanding of the framework's foundational principles. Section~\ref{sec_experimental_setting} explains the experimental setup and analyzes the experimental results.

In this study, we primarily undertake the following innovations in expending ML-compass:
\begin{itemize}
    \item Involving more metrics in model utility evaluation.
    \item Introducing mutant methods in the model utility evaluation to simulate real-world scenarios.
    \item Incorporating black-box transfer attacks in the robustness evaluation to reveal potential model vulnerabilities in practice.
    \item Integrating more interpretability methods in the interpretability evaluation, using Insertion \& Deletion Score as a metric to quantitatively assess model interpretability.
    \item To enable neuron analysis, we employed additional pruning algorithms.
    \item Introducing a comprehensive evaluation, utilizing radar charts for a more intuitive display of model testing results, facilitating user selection of suitable models.
    \item Exploring additional testing possibilities for models by combining pruning algorithms with the robustness and interpretability analysis.
\end{itemize}

\begin{table*}[t]
\centering
\caption{Existing testing tools for DLS assessment} 
\label{tab:checklist}
\resizebox{\textwidth}{!}{%
\begin{threeparttable}
\begin{tabular}{@{}lccccccc@{}}
\toprule
 & \textbf{Basic Metric} & \textbf{Mutants} & \textbf{Neural Analysis} & \makecell{\textbf{Robustness Analysis}\\\textbf{(white-box)}} &  \makecell{\textbf{Robustness Analysis}\\\textbf{(black-box)}} & \textbf{Interpretability} & \textbf{Multi-Model} \\ 
\midrule
DeepXplore~\cite{pei2017deepxplore} & \pie{360} & \pie{0} & \pie{360} & \pie{360} & \pie{0} & \pie{0} & \pie{0} \\
NeuronFair~\cite{zheng2022neuronfair} & \pie{360} & \pie{0} & \pie{0} & \pie{0} & \pie{0} & \pie{360} & \pie{0} \\
DeepGauge~\cite{ma2018deepgauge} & \pie{360} & \pie{0} & \pie{0} & \pie{0} & \pie{0} & \pie{0} & \pie{0} \\
DeepTest~\cite{tian2018deeptest} & \pie{360} & \pie{0} & \pie{0} & \pie{0} & \pie{0} & \pie{0} & \pie{0} \\
DeepMutation~\cite{ma2018deepmutation} & \pie{360} & \pie{360} & \pie{0} & \pie{0} & \pie{0} & \pie{0} & \pie{0} \\
DeepMutation++~\cite{hu2019deepmutation++} & \pie{360} & \pie{360} & \pie{0} & \pie{360} & \pie{0} & \pie{0} & \pie{0} \\
InterpretDL~\cite{li2022interpretdl} & \pie{0} & \pie{0} & \pie{0} & \pie{0} & \pie{0} & \pie{360} & \pie{0} \\ 
\midrule
Ours (AI-Compass) & \pie{360} & \pie{360} & \pie{360} & \pie{360} & \pie{360} & \pie{360} & \pie{360} \\ 
\bottomrule
\end{tabular}%
\begin{tablenotes}
\item \textbf{\pie{360}}: The test tool has the function; \textbf{\pie{0}}: The test tool does not have the function.
\end{tablenotes}
\end{threeparttable}
}
\end{table*}

\section{Related work}\label{sec_related_work}

In this section, we introduce the existing testing tools for assessing deep learning models. Based on their functionalities, existing DLS testing tools could be categorized into three groups, focusing on adversarial robustness, model interpretability, and neuron analysis, separately.

\noindent \textbf{Testing tools for adversarial robustness.}
Adversarial attacks refer to malicious attempts by adversaries to introduce subtle yet meaningful perturbations to input data, with the aim of inducing misclassification or erroneous predictions from the model. Currently, mainstream adversarial attacks can be categorized into white-box attacks and black-box attacks. In a white-box setting, relevant information such as the structure and parameters of the target model are transparent. Leveraging this characteristic, white-box attack algorithms can generate high-quality adversarial examples to assess the target model's resilience against various types of attacks. Therefore, white-box attacks serve as an ideal approach to evaluate the robustness of models against adversarial attacks. By continuously challenging and attacking models, researchers and software developers can uncover and address hidden flaws, thereby contributing to the construction of more secure and reliable DLS that safeguard the security of users and data.
The Adversarial Robustness Toolbox (ART)~\cite{nicolae2018adversarial} is an open-source testing tool dedicated to evaluating and enhancing the robustness of deep learning models. It offers a range of white-box attack algorithms and defense mechanisms tailored for deep learning models, such as FGSM~\cite{goodfellow2014explaining}, DeepFool~\cite{moosavi2016deepfool}, and C\&W~\cite{carlini2017towards}, which are of significant relevance for assessing the robustness of systems. TextAttack~\cite{morris2020textattack}, as a testing tool focused on adversarial example generation and model robustness testing for natural language processing tasks, is suitable for white-box attack algorithms like HotFlip~\cite{ebrahimi2017hotflip} and TextFooler~\cite{jin2020bert}, providing support for security testing in text-based applications.

Compared to the white-box conditions, model information in the black-box setting is difficult to obtain. Furthermore, black-box attack algorithms can be used to simulate real-world security threats and exploit scenarios, which is crucial for enhancing model robustness. 
Foolbox~\cite{rauber2017foolbox} is a Python library for generating adversarial samples and evaluating models, which supports various black-box attack algorithms and demonstrates excellent performance across multiple deep learning frameworks such as PyTorch, Keras, and TensorFlow. TextBugger~\cite{li2018textbugger} is a black-box adversarial sample generation framework specifically designed for text classification tasks. It can be employed to assess the robustness of deep learning models in text-related tasks. However, the aforementioned DLS testing tools are limited in their applicability as they are designed for specific environments (white-box or black-box) for adversarial robustness evaluation. They do not constitute a universal testing tool and are incapable of effectively assessing model interpretability.



\noindent \textbf{Testing tools for Model Interpretability.}
Performing interpretability analysis on models is an effective approach to understanding the process by which models generate predictions for different inputs. Moreover, interpretability analysis of models serves as a tool for elucidating the reasons behind errors encountered during DLS testing, thereby enhancing the trustworthiness of models and constituting a vital component of Explainable AI (XAI) research. Presently, several DL testing tools have been developed to elucidate the internal workings and decision processes of DLS, aiming to ensure system quality and reliability.
InterpretDL~\cite{li2022interpretdl} provides a range of functionalities, including feature importance analysis, sample explanations, and model visualization, enabling users to analyze model predictions and gain insights into the underlying patterns and information embedded within the model. NeuronFair~\cite{zheng2022neuronfair}addresses reliability and fairness concerns that may arise when applying DNNs in sensitive domains, which serves as a fairness testing framework for building fairer and more trustworthy DLS. However, InterpretDL and NeuronFair often rely on specific DL frameworks, which can be limiting for researchers and practitioners using customized or less common frameworks. In addition, the interpretability of DL models is a complex issue. InterpretDL and NeuronFair do not provide a complete explanation for every decision made by the model, as the effectiveness of model interpretation may be constrained by model complexity. Similarly, the aforementioned methods solely assess the interpretability of DLS, lacking consideration for adversarial robustness.





\noindent \textbf{Testing tools for neuron analysis.}
A testing tool providing neural analysis utilizes the properties of neurons, including the parameter of each neuron or the performance of each neuron's output under the specific testing input. Many testing tools primarily focus on this aspect. DeepXplore~\cite{pei2017deepxplore} is the first testing tool to propose neural coverage and use the joint optimization method with gradient ascent to generate testing examples. DeepGauge~\cite{ma2018deepgauge} uses neuron states to supervise the purpose of multi-granularity testing coverage. DeepTest~\cite{tian2018deeptest} uses transformation operations to get higher neural coverage under the specific field of DNN-driven autonomous cars. 
Although neural coverage is a widely utilized criterion among testing tools, several studies~\cite{dong2019there,harel2020neuron} have revealed that relying solely on neural coverage can lead to the generation of misleading testing examples. Overemphasizing neural coverage may result in a limited number of test inputs, potentially overlooking defects in DLS. Based on the premise, Jin \etal~\cite{jin2023excitement} use shapley value to define excitable neural, which can be regarded as other types of neural analysis. In our research endeavor, we harness the pruning property inherent in DLS, which entails the removal of extraneous neurons from the neural networks through neural analysis. By employing this approach, we aim to discern the intricate relationship between the network's robustness and the presence of indispensable neurons.

\section{Preliminaries}\label{sec_preliminaries}
In this section, we introduce preliminaries of adversarial attacks, interpretability methods, and neuron pruning methods for evaluating adversarial robustness, model explainability, and model's neuron analysis.


\subsection{Adversarial attacks}
During the development of ML and DL, DNNs have been proved to have state-of-the-art results in massive fields such as image classification~\cite{li2020learning}, speech recognition~\cite{seltzer2013investigation}, natural language processing~\cite{yin2017comparative}, and recommendation systems~\cite{lian2018xdeepfm}. As a multiple-layer unsupervised neural network, the output of DNN is available by layer-to-layer mapping, which can effectively extract the hidden features from the input space and achieve outstanding performance beyond human. Besides, several optimal training methods such as dropout regularization~\cite{jindal2016learning} and mini-batching~\cite{dixon2017classification} serve to reduce the computation cost of DNNs, allowing the model to have a high prediction accuracy and a fast convergence speed. However, the complex decision boundary of DNNs raises a threat in software quality~\cite{guan2020analysis}. Adversarial samples with human-add perturbations as well as noises existed in real-world enable an issue of incorrect model predictions~\cite{biggio2013evasion}. For example, a DL software system is easily fooled in an image classification task due to its vulnerability towards adversarial samples in the pixel space~\cite{su2019one}. The instability of DNNs in the face of adversarial attacks will cause serious security problems, especially in some practical applications that require low false-positive rates (\eg autonomous driving~\cite{deng2020analysis} and cancer detection~\cite{huq2020analysis}). It is thus necessary and urgent to explore a deep learning testing tool that can measure the adversarial robustness of DLS.

Nowadays adversarial algorithms are a major approach to test the robustness of models under attack because of the ability to generate promising adversarial samples. Generally, according to the model information that can be accessed, adversarial algorithms can be divided into two categories: white-box attacks~\cite{goodfellow2014explaining,kurakin2018adversarial,madry2017towards,carlini2017towards,dong2017discovering,dong2019evading,xiao2018generating} and black-box attacks~\cite{li2020qeba,chen2017zoo,lin2019nesterov,xie2019improving,gao2020patch,zhang2022improving,long2022frequency,jin2023danaa,10.1145/3583780.3614927}. In the white-box environment, the model information(\eg parameters and structure) can be visited by the attacker. On the contrary, in the black-box environment it is difficult to obtain model details. 

\noindent \textbf{White-box adversarial attacks.}
Gradient-based white-box attacks aim to apply advanced gradient operations to increase the success rate of attack. The FGSM~\cite{goodfellow2014explaining} algorithm and some of its derivatives such as I-FGSM~\cite{kurakin2018adversarial}, MI-FGSM~\cite{dong2017discovering}, TI-FGSM~\cite{dong2019evading}, SINI-FGSM~\cite{lin2019nesterov}, etc. have been proved to have an excellent success rate in the white box model. PGD~\cite{madry2017towards} and C\&W~\cite{carlini2017towards} algorithms focus on restricting perturbations or using special mathematical constraints to improve the robustness of adversarial examples. In addition, AdvGAN~\cite{xiao2018generating} generates adversarial samples by learning a generator network instead of perturbing input samples directly. 

\noindent \textbf{Black-box adversarial attacks.}
As query-based algorithms in black-box attacks, QEBA~\cite{li2020qeba} and ZOO~\cite{chen2017zoo} rely on small batches of queries to obtain model information to train adversarial samples. While SSA~\cite{long2022frequency}, DIM~\cite{xie2019improving}, PIM~\cite{gao2020patch}, RAP~\cite{qin2022boosting} and NAA~\cite{zhang2022improving} are trained on a surrogate model to test the effectiveness of adversarial samples when transferred into the target model.



\subsection{Model interpretability}

Recently, DNNs remain to be difficult to be interpreted due to the complex hidden layer parameters and incomprehensible nonlinear structure. It is currently unclear how deep models interpret the relationship between their inputs and outputs. Exploring the ambiguous decision-making process of DNNs is an important task in Explainable Artificial Intelligence (XAI) research. For DL software quality testing, a trustworthy system not only needs high accuracy, but also requires to have easy-to-interpret properties in the process of obtaining the results~\cite{adadi2018peeking}. Therefore, the verification of model interpretability is an important factor for in-depth exploration of the eligibility of DL systems.

To get the corresponding information for the model features and predictions, local approximation methods and gradient based methods are two common directions towards interpreting DNNs. The former attempts to obtain an approximate explanation of the complex target model through a relatively simple and interpretable model, while the latter aims to use the gradient information of the model to obtain the specific relationship between the input features and the outputs. 

\noindent \textbf{Local approximation methods.~}
Linear models and decision tree models are widely used in local approximation methods due to the high interpretability of these models~\cite{ribeiro2016should,shrikumar2017learning,lundberg2017unified}. Other approximation methods add perturbations to training data to obtain the most sensitive part of the inputs with respect to the model outputs~\cite{fong2017interpretable,datta2016algorithmic,li2016understanding}.

\noindent \textbf{Gradient based methods.~}
As two early gradient based methods, Grad-CAM~\cite{selvaraju2017grad} and Score-CAM~\cite{wang2020score} are both class activation mapping (CAM) based methods which aim to explain the relationship between gradient information and intermediate layers of DNN feature maps. Saliency Map~\cite{patra2020incremental} (SM) applies gradient directly to obtain the visualisation of the particular features with respect to the model outputs. Guided Backpropagation~\cite{springenberg2014striving} uses non-negative gradients of the model to get the desired explanation. However, Guided Backpropagation is poorly interpretable for features in negative gradient directions. In addition, only using gradient information is limited for current deep models with increasingly complex structures and diverse application scenarios. For example, SM suffers from gradient saturation and interpretation distortions caused by some noise or changes in external conditions.

To solve the misinterpretation of gradients in specific regions existing in earlier gradient analysis methods, the IG~\cite{sundararajan2017axiomatic} attribution algorithm first extends the original simple gradient calculation into a linear gradient integral from baseline features to input features, improving the interpretability of the model. Introducing prior knowledge as a prior probability distribution for feature attribution, EG~\cite{erion2021improving} has obtained further interpretability improvement on the basis of IG. BIG~\cite{wang2021robust} is firstly proposed to use adversarial attack to
determine suitable decision boundaries and apply the attribution method based on IG to find the exact information leading to these decision boundaries. AGI~\cite{pan2021explaining} noted that the IG method must seek a specific reference point in the attribution path as a starting point for iteration. In different models, the selection of reference points is complex and unique, which is not conducive to the generalization of IG. Therefore, AGI uses the gradient information of the adversarial sample to integrate along the path with the steepest gradient, so that the contribution of all input features can be calculated without selecting a reference point.

\subsection{Algorithm pruning}
The initial purpose of the pruning algorithm was to reduce the computational cost of the DL system. The pruning algorithm means preserving valuable parameters in the model while removing redundant parameters~\cite{10.1145/3583780.3614889}. Some work~\cite{jordao2021effect,ye2019adversarial} proves that connections between model pruning and robustness exist. From another perspective, the pruning algorithm naturally determines which neuron undertakes the work of model decision-making. As many testing tools have claimed, neural coverage can help increase the quality of testing examples. In order to gain a comprehensive understanding of DLS, it is recommended to employ a pruning algorithm that imposes a stringent constraint on the neurons prior to conducting efficient robustness testing. By implementing pruning techniques, we can alleviate concerns regarding testing methodology bias and focus our efforts on identifying and eliminating redundant parameters.

It is important to note that pruning algorithms can be categorized into two types: those with fine-tuning~\cite{liu2018fine} and those without fine-tuning~\cite{xu2021efficient}. Our research specifically concentrates on the design of pruning algorithms without fine-tuning. This choice is motivated by the fact that fine-tuning alters the original parameters, even if it improves performance. Our objective in pruning is to selectively remove certain parameters from DLS while preserving others entirely.

In earlier studies, Hu \etal~\cite{hu2016network} proposed a pruning algorithm to eliminate neurons with zero activation. Subsequently, similar pruning algorithms have placed greater emphasis on the dynamic performance of DLS, such as OBD~\cite{lecun1989optimal}. OBD employs second-order performance estimation to assess the importance of each neuron. Additionally, Taylor~\cite{molchanov2019importance} utilizes Taylor expansion to estimate the contribution of individual neurons in decision-making. Greg-2~\cite{wang2020neural} applies regularization techniques to constrain the estimation of neuron importance, utilizing a clever method to obtain relative importance differences instead of directly calculating the Hessian matrix. OBD, Taylor, Greg-2, and ASL~\cite{retsinas2020weight} are capable of being executed without the need for fine-tuning, and our work will integrate these approaches.

\section{\tool Structure Overview}\label{sec_structure_overview}

\begin{figure*}[t]
\label{structure chart}
\centering
\includegraphics[width=0.95\linewidth]{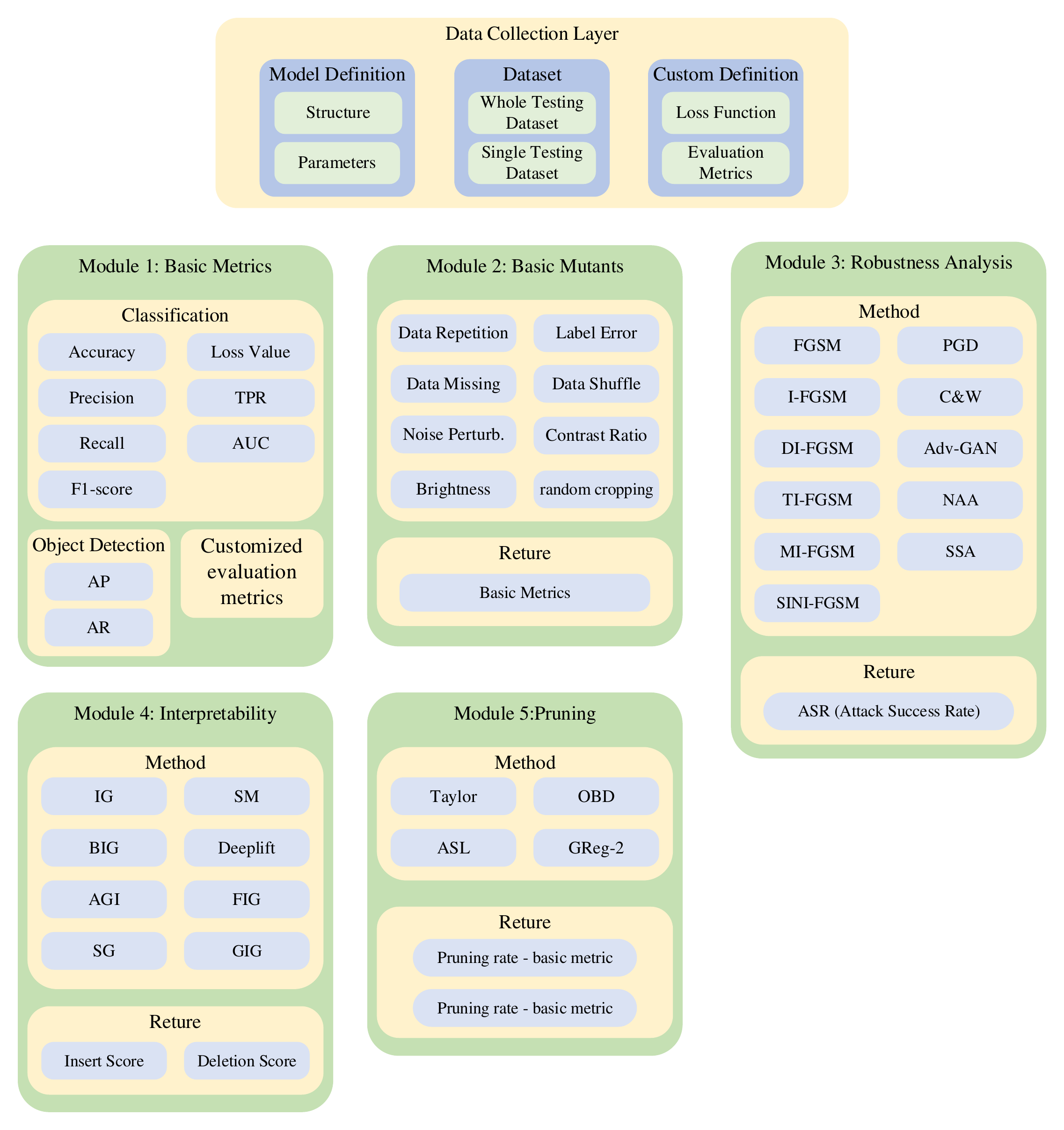}
\caption{An overview of \tool.}
\label{fig:structure}
\end{figure*}

In this section, we provide a general overview of our work, an all-in-one comprehensive and effective multi-modal testing tool for DLS. The main components of our framework are shown in Figure~\ref{fig:structure}. We aim to develop a framework that exhibits excellent testing performance in both image and text input data, catering to the needs of testing across different modalities. Specifically, we have designed five modules, namely Basic Metrics, Basic Mutants, Robustness Analysis, Interpretability, and Neuron Analysis, to comprehensively test DLS. Within each module, we employ appropriate evaluation metrics to obtain the most reasonable assessment results for the corresponding module. Moreover, we introduce pruning techniques to analyze the redundancy levels of model neurons and investigate the potential alterations in individual modules. This serves as a crucial foundation for optimizing model structure.

It is noteworthy that our framework is an all-in-one solution, serving as a comprehensive, multifunctional, and customizable DLS testing tool. During the preparation stage, diverse DLS undergo two initial collection layers to collect model and dataset information for customizable testing, followed by comprehensive assessment in multiple modules. The visualization of results and the generation of testing reports are provided for researchers or software developers to evaluate the quality of the systems. Users can also customize the testing methods, evaluation techniques, and corresponding metrics based on their specific systems and testing requirements. This ensures the adaptability and flexibility of the testing tool. Table~\ref{tab:requirements} illustrates an overview of the data requirements and supported tasks for each module.


\begin{table*}[t]
\centering
\caption{Data requirements and supported tasks for each module in the assessment layer of ML-Compass.}\label{tab:requirements}
\resizebox{\linewidth}{!}{
\begin{threeparttable}
\begin{tabular}{@{}c|c|ccc|ccc@{}}
\toprule
\textbf{Module} 
& \textbf{Submodule} 
& \begin{tabular}[c]{@{}c@{}}\textbf{Single}\\\textbf{Test Dataset}\end{tabular}
& \begin{tabular}[c]{@{}c@{}}\textbf{Whole}\\\textbf{Test Dataset}\end{tabular}
& \begin{tabular}[c]{@{}c@{}}\textbf{Train}\\ \textbf{Dataset}\end{tabular} 
& \begin{tabular}[c]{@{}c@{}}\textbf{Image}\\ \textbf{Classification}\end{tabular}  
& \begin{tabular}[c]{@{}c@{}}\textbf{Text}\\ \textbf{Classification}\end{tabular} 
& \begin{tabular}[c]{@{}c@{}}\textbf{Object}\\ \textbf{Detection}\end{tabular}  \\ 
\midrule
\multirow{9}{*}{Basic Metrics} & Accuracy & \pie{0} & \pie{360} & \pie{0} & \pie{360} & \pie{360} & \pie{0} \\
 & Loss Value & \pie{0} & \pie{360} & \pie{0} & \pie{360} & \pie{360} & \pie{0} \\
 & Precision & \pie{0} & \pie{360} & \pie{0} & \pie{360} & \pie{360} & \pie{0} \\
 & TPR & \pie{0} & \pie{360} & \pie{0} & \pie{360} & \pie{360} & \pie{0} \\
 & Recall & \pie{0} & \pie{360} & \pie{0} & \pie{360} & \pie{360} & \pie{0} \\
 & AUC & \pie{0} & \pie{360} & \pie{0} & \pie{360} & \pie{360} & \pie{0} \\
 & F1-score & \pie{0} & \pie{360} & \pie{0} & \pie{360} & \pie{360} & \pie{0} \\
 & AP & \pie{0} & \pie{360} & \pie{0} & \pie{360} & \pie{360} & \pie{360} \\
 & AR & \pie{0} & \pie{360} & \pie{0} & \pie{360} & \pie{360} & \pie{360} \\ 
\midrule
\multirow{7}{*}{Basic Mutants} 
 & Label Error & \pie{360} & \pie{360} & \pie{0} & \pie{360} & \pie{360} & \pie{0} \\
 & Data Missing & \pie{360} & \pie{360} & \pie{0} & \pie{360} & \pie{360} & \pie{360} \\
 & Data Shuffle & \pie{360} & \pie{360} & \pie{0} & \pie{360} & \pie{360} & \pie{360} \\
 & Noise Perturb & \pie{360} & \pie{360} & \pie{0} & \pie{360} & \pie{360} & \pie{360} \\
 & Contrast Ratio & \pie{360} & \pie{360} & \pie{0} & \pie{360} & \pie{0} & \pie{360} \\
 & Brightness & \pie{360} & \pie{360} & \pie{0} & \pie{360} & \pie{0} & \pie{360} \\
 & Random Cropping & \pie{360} & \pie{360} & \pie{0} & \pie{360} & \pie{0} & \pie{360} \\ 
\midrule
\multirow{11}{*}{Robustness Analysis} & FGSM & \pie{360} & \pie{360} & \pie{0} & \pie{360} & \pie{360} & \pie{360} \\
 & I-FGSM & \pie{360} & \pie{360} & \pie{0} & \pie{360} & \pie{360} & \pie{360} \\
 & DI-FGSM & \pie{360} & \pie{360} & \pie{0} & \pie{360} & \pie{0} & \pie{360} \\
 & TI-FGSM & \pie{360} & \pie{360} & \pie{0} & \pie{360} & \pie{0} & \pie{360} \\
 & MI-FGSM & \pie{360} & \pie{360} & \pie{0} & \pie{360} & \pie{360} & \pie{360} \\
 & SINI-FGSM & \pie{360} & \pie{360} & \pie{0} & \pie{360} & \pie{360} & \pie{360} \\
 & PGD & \pie{360} & \pie{360} & \pie{0} & \pie{360} & \pie{360} & \pie{360} \\
 & C\&W & \pie{360} & \pie{360} & \pie{0} & \pie{360} & \pie{0} & \pie{0} \\
 & Adv-GAN & \pie{360} & \pie{360} & \pie{0} & \pie{360} & \pie{0} & \pie{0} \\
 & NAA & \pie{360} & \pie{360} & \pie{0} & \pie{360} & \pie{0} & \pie{360} \\
 & SAA & \pie{360} & \pie{360} & \pie{0} & \pie{360} & \pie{0} & \pie{360} \\ 
 \midrule
\multirow{8}{*}{Interpretability}
 & IG & \pie{360} & \pie{360} & \pie{0} & \pie{360} & \pie{0} & \pie{0} \\
 & BIG & \pie{360} & \pie{360} & \pie{0} & \pie{360} & \pie{0} & \pie{0} \\
 & AGI & \pie{360} & \pie{360} & \pie{0} & \pie{360} & \pie{0} & \pie{360} \\
 & SG & \pie{360} & \pie{360} & \pie{0} & \pie{360} & \pie{0} & \pie{0} \\
 & SM & \pie{360} & \pie{360} & \pie{0} & \pie{360} & \pie{0} & \pie{0} \\
 & Deeplift & \pie{360} & \pie{360} & \pie{0} & \pie{360} & \pie{0} & \pie{0} \\
 & FIG & \pie{360} & \pie{360} & \pie{0} & \pie{360} & \pie{0} & \pie{360} \\
 & GIG & \pie{360} & \pie{360} & \pie{360} & \pie{360} & \pie{0} & \pie{0} \\ 
\midrule
\multirow{4}{*}{Neuron analysis} 
 & Taylor & \pie{0} & \pie{360} & \pie{360} & \pie{360} & \pie{0} & \pie{0} \\
 & ASL & \pie{0} & \pie{360} & \pie{360} & \pie{360} & \pie{0} & \pie{0} \\
 & OBD & \pie{0} & \pie{360} & \pie{360} & \pie{360} & \pie{0} & \pie{0} \\
 & Greg-2 & \pie{0} & \pie{360} & \pie{360} & \pie{360} & \pie{0} & \pie{0} \\ 
\bottomrule
\end{tabular}
\begin{tablenotes}
\item \textbf{\pie{360}}: the dataset is required (either dataset will fulfill the requirement) or the task is supported; \textbf{\pie{0}}: the dataset is not required or the task is not supported.
\end{tablenotes}
\end{threeparttable}
}
\end{table*}

\noindent \textbf{Module 1 Basic Metrics.~} We employ various fundamental DL evaluation metrics to conduct preliminary testing of DLS. Specifically, for classification tasks, we evaluate the performance using metrics such as accuracy, precision, recall, and loss value. For object detection tasks, we utilize metrics such as average precision (AP) and average recall (AR) to assess the performance.

\noindent \textbf{Module 2 Basic Mutants.~} We apply various mutant methods such as label error, data repetition, data missing, and noise perturbation to test the specific performance of DLS. The evaluation metrics used in this module are the same as those in Module 1.

\noindent \textbf{Module 3 Robustness Analysis.~} We employ white-box attack methods such as FGSM~\cite{goodfellow2014explaining}, PGD~\cite{madry2017towards}, and C\&W~\cite{carlini2017towards}, as well as black-box attack methods including MI-FGSM~\cite{dong2017discovering}, NAA~\cite{zhang2022improving} and SSA~\cite{long2022frequency} to test the robustness of the model against adversarial attacks in both the strongest attack setting and the simulated real-world environment. This enables us to identify and rectify vulnerabilities or loopholes hidden within the system. To effectively evaluate the model's robustness against adversarial attacks, we utilize the attack success rate (ASR) as a benchmark metric in Module 3.

\noindent \textbf{Module 4 Interpretability.~} In order to understand the intrinsic connection between model outputs and inputs and provide better insights into the decision process at the decision boundary, we employ algorithms such as IG~\cite{sundararajan2017axiomatic}, BIG~\cite{wang2021robust}, GIG~\cite{kapishnikov2021guided}, and AGI~\cite{pan2021explaining} to obtain model interpretation results. It is worth noting that in Module 4, we employ insertion score and deletion score~\cite{petsiuk2018rise} to evaluate the effectiveness of the model's interpretability.

\noindent \textbf{Module 5 Neuron Analysis.~} Previous works have demonstrated the remarkable effectiveness of Neuron Analysis in fault detection and localization, optimized test sample generation, and understanding model complexity in DLS testing. Specifically, Neuron Analysis reveals the proportion of effectively utilized neurons, aiding in the discovery of potentially overlooked hidden behaviors or boundary cases, and guiding decisions on model optimization, compression, or pruning. Inspired by this, we integrate state-of-the-art pruning algorithms (\eg OBD~\cite{lecun1989optimal}, Greg-2~\cite{wang2020neural}, and ASL~\cite{retsinas2020weight}) into Neuron Analysis Module. These algorithms help analyze the redundancy of model neurons to explore the deep performance of the model and the potential relationships between modules in combination with robustness and interpretability analysis. The Pruning rate serves as the fundamental metric in this module.

\section{Methodology}\label{sec_methodology}
In this section, we provide a comprehensive technical description of our \tool. In the first step, we define and explain the concepts of adversarial attacks and the two attribution axioms. In the second step, we describe the representative methods or algorithms employed in each module. Specifically, in addition to conducting basic metric tests and mutation tests, we comprehensively evaluate model's adversarial robustness by integrating various white-box and black-box adversarial attack algorithms. This approach addresses the limitations of existing testing tools, which are typically confined to either white-box or black-box environments. To provide the most effective explanations for the model's behavior, we incorporate several state-of-the-art attribution algorithms to generate explainable reports for the model. It is worth noting that, for the first time, we have selected several pruning algorithms without fine-tuning as a means of neuron analysis. By testing the pruned model in conjunction with other modules, we analyze the redundancy of the model's neurons and assess its deep performance based on changes in adversarial robustness and model interpretability. For example, if there is no significant change in adversarial robustness or model interpretability after pruning, it can be inferred that the original model contains redundant parameters and has room for further optimization.

\subsection{Problem Definition and Invariance Theory}
\subsubsection{Problem Definition of Adversarial Attack}
Formally, suppose we have a deep neural network $N:$ $\mathbb{R}^n \to \mathbb{R}^c$ and original sample $x \in R^n$. Adversarial attack methods, in this context, are designed to uncover a perturbation denoted as $\Delta x$. This perturbation is intended to be added to the original sample $x$, thereby generating a manipulated input sample denoted as $x' = x + \Delta x$. The overarching objective of the optimization process, in this scenario, is to satisfy the following conditions
\begin{equation}
\label{eqattack}
\min_{x'} D(x, x') \quad \text{subject to} \quad N(x) \neq N(x')
\end{equation}
We note that for the classification task, $N(x)$ and $N(x')$ should satisfy the constraint as in Equation~\ref{eqattack}, \ie $N(x) \neq N(x')$. And for the regression task, the constraint can be defined as $N(x) - N(x') \geq \epsilon $, depending on the application.

\subsubsection{Sensitivity and Implementation Invariance}\label{axiom}
In the context of Integrated Gradients~\cite{sundararajan2017axiomatic} (IG), two crucial concepts are introduced to address the requirements for explaining DNNs: sensitivity and implementation invariance. We believe that these two axioms are of paramount importance in the process of model interpretation.

\noindent \textbf{Sensitivity.~} Sensitivity pertains to the degree of responsiveness exhibited by an attribution method towards slight perturbations in the input data. In particular, when comparing inputs and baselines that only vary in a single feature but yield different predictions, the attribution method should assign non-zero attributions to the differing features.

\noindent \textbf{Implementation Invariance.~} Implementation Invariance refers to the attribute of an attribution method that remains unaffected by variations in the implementation of a deep neural network. In formal terms, two networks are considered functionally equivalent if their outputs are equal for all inputs, irrespective of having significantly different implementations. To satisfy Implementation Invariance, attribution methods should consistently produce identical attributions for two functionally equivalent networks.

\subsection{Assessment algorithms}

\subsubsection{Basic \& mutant testing}
Typically, as a basic module, the fundamental metrics testing exhibits simplicity and provides rapid feedback on test results. We utilize evaluation metrics, as demonstrated in Module 1 of Section~\ref{sec_structure_overview}, to obtain a basic quality estimation of the system. However, recognizing that baseline testing fails to reflect the system's performance under unexpected circumstances, we employ mutation methods in Module 2 to simulate more diverse data distributions and noise scenarios. We consider this as one of the criteria for measuring system stability. 

Specifically, in Module 1 and Module 2, it is mandated that the user provides the requisite dataloader for testing purposes. However, a distinction arises between the two modules in terms of the model provision. Module 1 necessitates the user to supply the pre-trained model of their choice, whereas Module 2 does not impose this requirement. Moreover, Module 1 yields the performance metrics values alongside a classification report that is saved as a CSV file. Conversely, Module 2 delivers the dataset after mutant.

\subsubsection{Adversarial robustness}
As an extension of traditional deep learning system testing, we acknowledge the significance of the system's robustness against adversarial attacks as an indispensable aspect of quality assessment, particularly in the context of privacy and security concerns. In this section, we provide a detailed description of the principles behind the adversarial attack algorithms in Module 3. We present a series of representative algorithms in both white-box and black-box environments. Notably, as shown in Table~\ref{tab:checklist}, we innovatively adopt transfer-based attack approaches in the black-box setting to evaluate the model's performance under simulated real-world conditions, which has yet to be proposed in current relevant testing frameworks.

The use of white-box attacks allows for the exploration of a model's robustness when its structure or parameters are potentially exposed, representing a consideration of the worst-case scenario in real-world applications. However, the probability of model parameters or structure leakage in real-life situations is relatively low. Therefore, black-box attacks are more in line with real-world scenarios, where attackers utilize surrogate models to launch attacks on the target model. As a result, we have incorporated a black-box attack testing module, which provides a better evaluation of the model's robustness when its information is not leaked. In general, users have the flexibility to choose whether to provide white-box information, including the model, test sample data, and corresponding labels, or to use black-box models to assess transferability. This choice depends on the specific testing requirements.

\noindent \textbf{FGSM.~} The Fast Gradient Sign Method (FGSM) is a commonly used white-box adversarial attack algorithm used to generate adversarial samples to deceive DNN models. The formula for FGSM is as follows:
\begin{equation}
    x'=x+\epsilon \cdot sign(\nabla_x J(\theta ,x,y))
\end{equation}
where $x'$ is the generated adversarial sample, $\epsilon$ is the hyperparameter that controls the size of the perturbation, $\nabla_x J(\theta, x, y)$ is the gradient of the loss function $J$ with respect to the input $x$, and $\text{sign}$ denotes the sign that takes the gradient. 

\noindent \textbf{PGD.~} Projected Gradient Descent (PGD) is an iterative adversarial attack algorithm which aims to gradually approach an adversarial example by applying small perturbations to the input sample based on gradient information in each iteration. The formulation of the PGD algorithm is as follows:
\begin{equation}
    x'=\Pi _{x+S}(x'+\alpha \cdot sign(\nabla_x J(\theta ,x',y)))
\end{equation}
Whereas, $\alpha$ represents the learning rate that controls the step size for each iteration. The $\Pi _{x+S}$ denotes the projection operation that restricts the perturbed input $x'$ to the valid range defined by $x+S$. $\nabla_x J(\theta, x', y)$ corresponds to the gradient of the loss function $J$ with respect to the input $x'$; $\text{sign}$ signifies taking the sign of the gradient.

\noindent \textbf{MI-FGSM.~} The Momentum Iterative Fast Gradient Sign Method (MI-FGSM) builds upon the FGSM algorithm by introducing the concept of momentum, which enhances the effectiveness and stability of attacks, particularly in the context of black-box transfer-based attacks.
\begin{equation}
    g \leftarrow \mu \cdot g + \frac{\nabla_x J(\theta, x', y)}{|\nabla_x J(\theta, x', y)|}
\end{equation}
\begin{equation}
    x' \leftarrow x' + \alpha \cdot \text{sign}(g)
\end{equation}
In each iteration, the momentum term $g$ accumulates gradient information and decays based on the momentum factor $\mu$. The introduction of momentum helps maintain a certain level of consistency in the gradient direction during the attack, thereby improving the success rate and stability of transfer-based attacks.

\subsubsection{Model interpretability}
Currently, the majority of testing tools utilize interpretability methods that focus on specific target class feature maps, such as Grad-CAM~\cite{selvaraju2017grad}, lacking a unified and systematic axiomatic discussion. Attribution algorithms based on Integrated Gradients (IG) introduced for the first time two axioms, Sensitivity and Implementation Invariance, which systematically establish a one-to-one correspondence between model outputs and inputs and provide explanations for features that are overlooked by traditional interpretability algorithms. A series of algorithms, including BIG and AGI, are dedicated to optimizing the potential drawbacks of IG to further enhance the accuracy of attributions. 

By comparing various interpretability methods, users can assess the model's interpretability capability. Specifically, by evaluating multiple instances of the same interpretability method, a model demonstrating superior evaluation metrics indicates better interpretability. This signifies a higher level of trustworthiness in the model. In this section, we primarily introduce attribution algorithms represented by IG and its variants to conduct a generic assessment of model interpretability.

\noindent \textbf{IG.~} As mentioned in ~\ref{axiom}, IG proposed two axiomatic criteria: \textit{Sensitivity} and \textit{Implementation Invariance}. By carefully selecting reference points as anchors along a linear integration path, IG effectively integrates the continuous gradients to determine the attribution of individual input features. The formula of IG is expressed in Equation~\ref{eqig}.
\begin{equation}
\label{eqig}
IG_{j}(x)=(x_{j}-x_{j}')\times\int_{\alpha=0}^{1} \frac{\partial F(x'+\alpha \times(x-x'))}{\partial x_{j}}\mathrm{d}\alpha  
\end{equation}
where $j$ denotes the $j$-th input feature, $\frac{\partial F(x'+\alpha \times(x-x'))}{\partial x_{j}}$ is the gradient of model $F$ w.r.t input feature $x_{j}$. $x_{j}'$ represents the reference input feature.

\noindent \textbf{BIG.~} 
Through investigating improved baseline selection techniques in comparison to IG, the Boundary-based Integrated Gradient (BIG) method introduces boundary search to achieve more precise attribution outcomes. Considering a deep learning network $F$, the Integrated Gradient $g_{IG}$, and an input feature $x$, the formula is expressed as Equation~\ref{eqbig}.
\begin{equation}
\label{eqbig}
B_{\mathrm{IG}}(\mathbf{x}):=g_{\mathrm{IG}}\left(\mathbf{x};\mathbf{x}^{\prime}\right) 
\end{equation}
where $x'$ is the nearest adversarial example to $x$, \ie $c = F(x) \ne F(x’) $ and $\forall \mathbf{x}_m \cdot\left\|\mathbf{x}_m-\mathbf{x}\right\|<\left\|\mathbf{x}^{\prime}-\mathbf{x}\right\| \rightarrow F(\mathbf{x})= F(\mathbf{x_m})$.

\noindent \textbf{AGI.~} The Adversarial Gradient Integration (AGI) method aims to identify the steepest non-linear ascending path from the adversarial example $x_i'$ to $x$, eliminating the requirement for reference points along the path, unlike IG. The formula is defined as Equation \ref{eqagi} in the following:
\begin{equation}
\label{eqagi}
AGI_{j}(x)=AGI_{j-1}(x)-\nabla_{x_j}f^{t}(x)\cdot \epsilon \cdot sign(\frac{\nabla_{x_j}f^{i}(x)}{\left | \nabla_{x_j}f^{i}(x) \right | } )
\end{equation}
$\nabla_{x_j} f^t(x)$ represents the gradient of the output value $f^t(x)$ w.r.t the $j$-th input feature $x$, where $t$ denotes the true class label. Similarly, $\nabla_{x_j} f^i(x)$ represents the gradient corresponding to the false class label $i$. The step size is denoted by $\epsilon$. The integration process continues along the path until $argmax_l f^l(x) = i$, indicating that the attribution integration stops when the predicted class label becomes $i$.

To conduct interpretability analysis, the user is required to provide the model, the test data along with its corresponding labels. By utilizing the aforementioned interpretability algorithms, the user can obtain the interpretability heat map or attribution map for a single image, as well as the average insertion and deletion scores for multiple test samples.

\subsubsection{Neuron analysis} 
In this section, we introduce for the first time the integration of pruning methods into our testing tool. By pruning model parameters at a certain proportion, we observe the performance changes. If the performance changes are small or within an acceptable range, it indicates that the pruned parameters are redundant. Hence, pruning algorithms can be used to evaluate model redundancy.

Furthermore, models with excessive redundancy are susceptible to the risk of overfitting. By employing different pruning rates and methods, we investigate the robustness and interpretability of the model under various conditions. This assists users in selecting models that better meet their requirements without the need for retraining, while minimizing significant performance changes.

We compare the changes in ASR and Insertion \& Deletion Score~\cite{pan2021explaining} before and after employing pruning algorithms, examining the impact of pruning on the modules of adversarial robustness and model interpretability. Furthermore, we proceed to introduce several representative pruning algorithms.

\noindent \textbf{OBD.~} The Optimal Brain Damage (OBD) algorithm employs a second-order expansion technique to estimate the performance of DLS by considering the impact of removing specific neurons.
\begin{equation}
    \delta E=\frac{1}{2} \sum_{i} h_{i i} \delta u_{i}^{2}
\end{equation}
where $h_{ii}$ represents the Hessian matrix and $\delta u_{i}^{2}$ denotes the value associated with the $i$-th neuron, OBD allows for the pruning of neurons with minimal perturbation to the overall error ($\delta E$). In OBD, the calculation of the Hessian matrix is approximated to streamline the process. Typically, neurons exhibiting lower $\delta E$ are considered non-essential for the proper functioning of DLS.

\noindent \textbf{Taylor.~} The Taylor algorithm leverages a combination of regularization techniques and Taylor expansion to estimate the significance of neurons within a neural network.
\begin{equation}
    \delta E(u)=(g_{m} \delta u_{i})^2
\end{equation}
where $\delta u_{i}$ is the value of $i$-th neuron, $g_m$ is the gradient value under the regularization. $\delta E$ is simplified to use the first-order expansion for computational optimization.

\noindent \textbf{Greg-2.~} Greg-2 algorithm takes into account the importance of neurons in a dynamic and relative context. It eliminates the need to compute the Hessian matrix during the pruning process by considering the relative relationships between neurons. For a more comprehensive understanding of Greg-2 and another algorithm called ASL, please refer to the references~\cite{wang2020neural} and~\cite{retsinas2020weight}.

In summary, the Pruning module requires the provision of the test model, train dataset, test dataset, and the desired pruning scale by the user. Upon completion, the pruned model is returned, along with performance metrics such as accuracy before and after the pruning process.


\section{Experiment Setup}\label{sec_experimental_setting}

\subsection{Experimental Environment}
The present tool is developed on PyTorch 1.11. All experiments conducted in this study are performed on a server running Ubuntu 20.04.4, equipped with AMD EPYC 7642 48-Core Processor, NVIDIA RTX3090 GPU, and 80GB RAM.

\subsection{Datasets} In this study, we employed several well-known datasets from the domains of image classification, object detection, and text classification. Specifically, for the image classification task, the CIFAR-100 dataset~\cite{krizhevsky2009learning} was utilized. The COCO dataset~\cite{lin2014microsoft} was employed for the object detection domain. As for the text classification task, we utilized the STT-2 dataset~\cite{socher2013recursive}.

\subsection{Models} In this experiment, in order to demonstrate the comprehensiveness and effectiveness of our testing framework, we conducted three different categories of tasks: image classification, object detection, and text classification, within each module. Additionally, for each task, we tested two different models to examine the performance differences between them. We use ResNet-50~\cite{he2016deep} and VGG-16~\cite{simonyan2014very} for image classification tasks, TextCNN~\cite{kim2014convolutional} and AB-LSTM~\cite{liu2019ab} for text classification tasks, Faster R-CNN~\cite{ren2015faster} and RetinaNet~\cite{lin2017focal} for object detection tasks.

\subsection{Metrics} In our evaluation part, in addition to the basic and commonly used metrics mentioned in Module 1, such as accuracy, recall, precision, etc., we have also included additional metrics in the extended modules to provide a more comprehensive assessment of DLS performance.

Regarding the evaluation of model robustness, we utilized the Attack Success Rate (ASR), which represents the proportion of successful adversarial samples in the total number of attack samples. Thus, it can be used to evaluate the performance of an attack method on a specific model.

In terms of evaluating model interpretability, we introduced the concept of interpretability analysis for individual data samples and multiple data samples. For the former, analysis is conducted by examining the heatmaps returned by the evaluation framework, which utilize different colors to assess the accuracy of the DLS system in capturing salient features. In the case of multiple samples, we employed the Insertion score and Deletion score~\cite{petsiuk2018rise} for evaluation. The Insertion score involves starting with an empty image and progressively adding pixels based on their attribution scores, beginning with the highest score and moving towards the lowest. Similarly, the Deletion score is obtained by iteratively removing pixels from the original image in descending order of their attribution scores.

In the neural analysis of DLS, we employed the pruning rate as our evaluation metric, which represents the proportion of parameters pruned from the model out of the total model parameters.

\subsection{Parameter Setting} In this experiment, apart from the pruning rate, all other parameter settings followed the default parameters specified in the original method. Users have the flexibility to customize these parameters for their subsequent usage. As for the pruning rate parameter, we set it to 0.35, 0.4, 0.45, and 0.5, respectively.

\subsection{Research questions} In our experimental study, we aim to investigate and address the following research questions:

\begin{itemize}
    \item \textbf{RQ1:} Does \tool effectively integrate each module so as to provide a comprehensive assessment of the model's performance?
    \item \textbf{RQ2:} In addition to image classification tasks, does \tool meet the test requirements under other modal tasks such as text classification and object detection? Does it overcome the shortcoming of ad-hoc in existing testing tools?
   
    \item \textbf{RQ3:} Combining the Adversarial Robustness and Model Interpretability modules, can \tool use the pruning method to evaluate model depth performance and give optimization recommendations?
\end{itemize}


\section{Experimental results}\label{sec_experimental_results}
\subsection{Answer to RQ1}
In this section, we performed basic utility evaluation, robustness evaluation, interpretability analysis and neuron analysis with pruning to verify which model in each module shows superior performance.

\begin{table*}[t]
\centering
\caption{The results of basic metrics and mutant testing.}
\label{tab:module1}
\resizebox{\textwidth}{!}{%
\begin{tabular}{@{}c|c|ccccccccccccc@{}}
\toprule
Model & Method & Accuracy & Loss Value & TPR & TNR & PPV & NPV & FPR & FNR & FDR & ROC\_AUC & Precision & Recall & F1-Score \\ 
\midrule
\multirow{8}{*}{ResNet-50}
 & Origin Image & 0.7929 & 3.8597 & 0.7929 & 0.9979 & 0.7936 & 0.9979 & 0.0021 & 0.2071 & 0.2064 & 0.9902 & 0.7936    & 0.7929 & 0.7922   \\
 & Label Error & 0.7128 & 3.9354 & 0.7133 & 0.9971 & 0.7137 & 0.9971 & 0.0029 & 0.2867 & 0.2863 & 0.9402 & 0.7137 & 0.7133 & 0.7121   \\
 & Data Missing & 0.5893 & 4.0795 & 0.5893 & 0.9959 & 0.6854 & 0.9959 & 0.0041 & 0.4107 & 0.3146 & 0.9608 & 0.6854 & 0.5893 & 0.6013   \\
 & Data Shuffle & 0.7929 & 3.8597 & 0.7929 & 0.9979 & 0.7936 & 0.9979 & 0.0021 & 0.2071 & 0.2064 & 0.9902 & 0.7936 & 0.7929 & 0.7922   \\
 & Noise Perturb & 0.3796 & 4.2737 & 0.3796 & 0.9937 & 0.6011 & 0.9937 & 0.0063 & 0.6204 & 0.3989 & 0.8871 & 0.6011 & 0.3796 & 0.4047   \\
 & Contrast Ratio & 0.7884 & 3.8649 & 0.7884 & 0.9979 & 0.7888 & 0.9979 & 0.0021 & 0.2116 & 0.2112 & 0.9898 & 0.7888 & 0.7884 & 0.7876   \\
 & Brightness & 0.7883 & 3.8646 & 0.7883 & 0.9979 & 0.789  & 0.9979 & 0.0021 & 0.2117 & 0.211  & 0.9897 & 0.789 & 0.7883 & 0.7875   \\
 & Random Cropping & 0.7391 & 3.9369 & 0.7391 & 0.9974 & 0.7441 & 0.9974 & 0.0026 & 0.2609 & 0.2559 & 0.9863 & 0.7441 & 0.7391 & 0.7388   \\
\midrule 
\multirow{8}{*}{VGG-16} & Origin Image & 0.7259 & 3.91 & 0.7259 & 0.9972 & 0.7275 & 0.9972 & 0.0028 & 0.2741 & 0.2725 & 0.9855 & 0.7275 & 0.7259 & 0.7254   \\
 & Label Error & 0.6525 & 3.9821 & 0.653  & 0.9965 & 0.6543 & 0.9965 & 0.0035 & 0.347 & 0.3457 & 0.9363 & 0.6543 & 0.653  & 0.652    \\
 & Data Missing & 0.5163 & 4.1185 & 0.5163 & 0.9951 & 0.5698 & 0.9951 & 0.0049 & 0.4837 & 0.4302 & 0.9486 & 0.5698 & 0.5163 & 0.5135   \\
 & Data Shuffle & 0.7259 & 3.91 & 0.7259 & 0.9972 & 0.7275 & 0.9972 & 0.0028 & 0.2741 & 0.2725 & 0.9855 & 0.7275 & 0.7259 & 0.7254   \\
 & Noise Perturb & 0.3966 & 4.237 & 0.3966 & 0.9939 & 0.5388 & 0.9939 & 0.0061 & 0.6034 & 0.4612 & 0.9024 & 0.5388 & 0.3966 & 0.4138   \\
 & Contrast Ratio & 0.7182 & 3.9164 & 0.7182 & 0.9972 & 0.7198 & 0.9972 & 0.0028 & 0.2818 & 0.2802 & 0.985 & 0.7198 & 0.7182 & 0.7175   \\
 & Brightness & 0.7223 & 3.9152 & 0.7223 & 0.9972 & 0.7236 & 0.9972 & 0.0028 & 0.2777 & 0.2764 & 0.9849 & 0.7236 & 0.7223 & 0.7217   \\
 & Random Cropping & 0.648 & 3.9879 & 0.648 & 0.9964 & 0.6614 & 0.9964 & 0.0036 & 0.352  & 0.3386 & 0.9768 & 0.6614 & 0.648 & 0.6494   \\
\bottomrule
\end{tabular}
}
\end{table*}

\subsubsection{Basic utility evaluation} As shown in Table~\ref{tab:module1}, considering all the metrics collectively, it can be observed that under various data processing methods such as the original dataset, label errors, missing data, and shuffled data, ResNet-50 slightly outperforms VGG-16 with higher accuracy and lower loss values in these scenarios, indicating its superior performance and generalization capability in handling such data.

However, under the data processing method involving noise perturbation, VGG-16 exhibits a slight advantage over ResNet-50. Although VGG-16 achieves a slightly higher accuracy compared to ResNet-50, the difference between the two models is not statistically significant.

Therefore, taking into account the performance across different data processing methods, it can be concluded that ResNet-50 and VGG-16 perform comparably overall, but in most cases, ResNet-50 demonstrates slightly better performance than VGG-16.

\begin{table*}[t]
\centering
\caption{Table of model robustness evaluation results, the data in the table are ASR, the bolded data are the results of white box attack, the unbolded data are the results of black box attack.}
\label{tab:module3}
\resizebox{0.8\textwidth}{!}{%
\begin{tabular}{@{}c|c|cccccc@{}}
\toprule
Test Model & Attack Method & ResNet-50 & VGG-16 & Inception-v3 & DenseNet-121 & GoogLeNet & MobileNet-v2 \\ 
\midrule
\multirow{12}{*}{ResNet-50}
& FGSM & 79.11\% & 58.34\% & 69.14\% & 68.74\% & 66.59\% & 57.54\% \\
 & I-FGSM & 99.97\% & 53.25\% & 75.42\% & 75.20\% & 66.01\% & 44.57\% \\
 & DI-FGSM & 99.72\% & 67.30\% & 84.06\% & 83.93\% & 79.46\% & 63.88\% \\
 & TI-FGSM & 86.49\% & 22.66\% & 36.56\% & 30.84\% & 30.48\% & 22.30\% \\
 & MI-FGSM & 99.87\% & 64.47\% & 80.98\% & 81.33\% & 75.23\% & 56.99\% \\
 & SINI-FGSM & 98.37\% & 60.63\% & 76.23\% & 74.98\% & 69.82\% & 57.78\% \\
 & PGD & 99.99\% & 53.31\% & 74.21\% & 74.37\% & 64.93\% & 44.08\% \\
 & C\&W & 80.50\% & 6.57\% & 13.42\% & 14.49\% & 9.27\% & 4.12\% \\
 & Adv-GAN & 87.83\% & 86.18\% & 83.86\% & 79.93\% & 81.83\% & 52.05\% \\
 & NAA & 91.65\% & 70.61\% & 84.06\% & 64.99\% & 75.53\% & 65.04\% \\
 & SSA & 97.00\% & 60.27\% & 93.38\% & 91.88\% & 80.19\% & 57.67\% \\
 & \textbf{Average} & \textbf{92.77\%} & \textbf{54.87\%} & \textbf{70.12\%} & \textbf{67.33\%} & \textbf{63.58\%} & \textbf{47.82\%} \\ \midrule
\multirow{12}{*}{VGG-16} & FGSM & 57.99\% & 75.81\% & 63.35\% & 62.25\% & 62.49\% & 54.52\% \\
 & I-FGSM & 44.04\% & 99.17\% & 59.93\% & 56.05\% & 55.45\% & 40.21\% \\
 & DI-FGSM & 69.64\% & 97.49\% & 76.09\% & 73.65\% & 75.17\% & 63.54\% \\
 & TI-FGSM & 41.76\% & 91.96\% & 43.40\% & 37.59\% & 39.79\% & 31.36\% \\
 & MI-FGSM & 60.87\% & 98.51\% & 71.18\% & 68.33\% & 67.85\% & 53.33\% \\
 & SINI-FGSM & 59.41\% & 94.88\% & 66.42\% & 62.93\% & 62.53\% & 54.60\% \\
 & PGD & 41.24\% & 99.16\% & 57.84\% & 52.51\% & 52.56\% & 35.77\% \\
 & C\&W & 6.71\% & 68.20\% & 7.38\% & 7.56\% & 7.56\% & 3.69\% \\
 & Adv-GAN & 40.26\% & 91.03\% & 71.01\% & 55.58\% & 79.19\% & 36.10\% \\
 & NAA & 62.27\% & 98.68\% & 74.33\% & 53.03\% & 67.52\% & 61.39\% \\
 & SSA & 57.37\% & 90.40\% & 84.63\% & 79.85\% & 69.88\% & 50.62\% \\
 & \textbf{Average} & \textbf{49.23\%} & \textbf{91.39\%} & \textbf{61.41\%} & \textbf{55.39\%} & \textbf{58.18\%} & \textbf{44.10\%} \\ 
\bottomrule
\end{tabular}%
}
\end{table*}

\subsubsection{Robustness evaluation}
In the experiment of this module, we employed two different types of attack methods: white-box attacks and black-box attacks. As shown in Table~\ref{tab:module3}, for white-box attacks, VGG-16 demonstrates a lower ASR compared to ResNet-50. Therefore, on the dataset used in this experiment, VGG-16 exhibits better robustness than ResNet-50. Considering the practical application scenarios of DLS, we incorporated transfer-based black-box attacks to simulate real-world testing conditions. In the black-box attacks, VGG-16 consistently achieves a lower ASR than ResNet-50, indicating it has better robustness than ResNet-50 in this particular task.

\begin{table}[t]
\centering
\caption{Insertion and Deletion Score of ResNet-50 and VGG-16}
\label{tab:module4}
\resizebox{\linewidth}{!}{%
\begin{tabular}{@{}c|c|cc@{}}
\toprule
Model & Method & Insertion Score & Deletion Score \\ 
\midrule
\multirow{10}{*}{ResNet-50}
& IG & 0.1136 & 0.0246 \\
 & BIG & 0.2272 & 0.042 \\
 & AGI & 0.3881 & 0.0463 \\
 & SG & 0.2352 & 0.0197 \\
 & SM & 0.1242 & 0.0332 \\
 & DeepLIFT & 0.1246 & 0.0256 \\
 & FIG & 0.0889 & 0.0314 \\
 & GIG & 0.1267 & 0.0186 \\
 & SaliencyMap & 0.2559 & 0.0479 \\
 & Average & 0.1872 & 0.0321 \\ \midrule
\multirow{10}{*}{VGG-16}
& IG & 0.0804 & 0.02 \\
 & BIG & 0.1828 & 0.0316 \\
 & AGI & 0.3428 & 0.0393 \\
 & SG & 0.1343 & 0.0162 \\
 & SM & 0.0834 & 0.0242 \\
 & DeepLIFT & 0.0956 & 0.0182 \\
 & FIG & 0.0682 & 0.0244 \\
 & GIG & 0.0906 & 0.0165 \\
 & SaliencyMap & 0.2279 & 0.0336 \\
 & Average & 0.1451 & 0.0249 \\ 
\bottomrule
\end{tabular}%
}
\end{table}

\begin{figure*}[t]
\centering
\subfigure[ResNet-50 Interpretability Result]{
\includegraphics[width=0.48\textwidth]{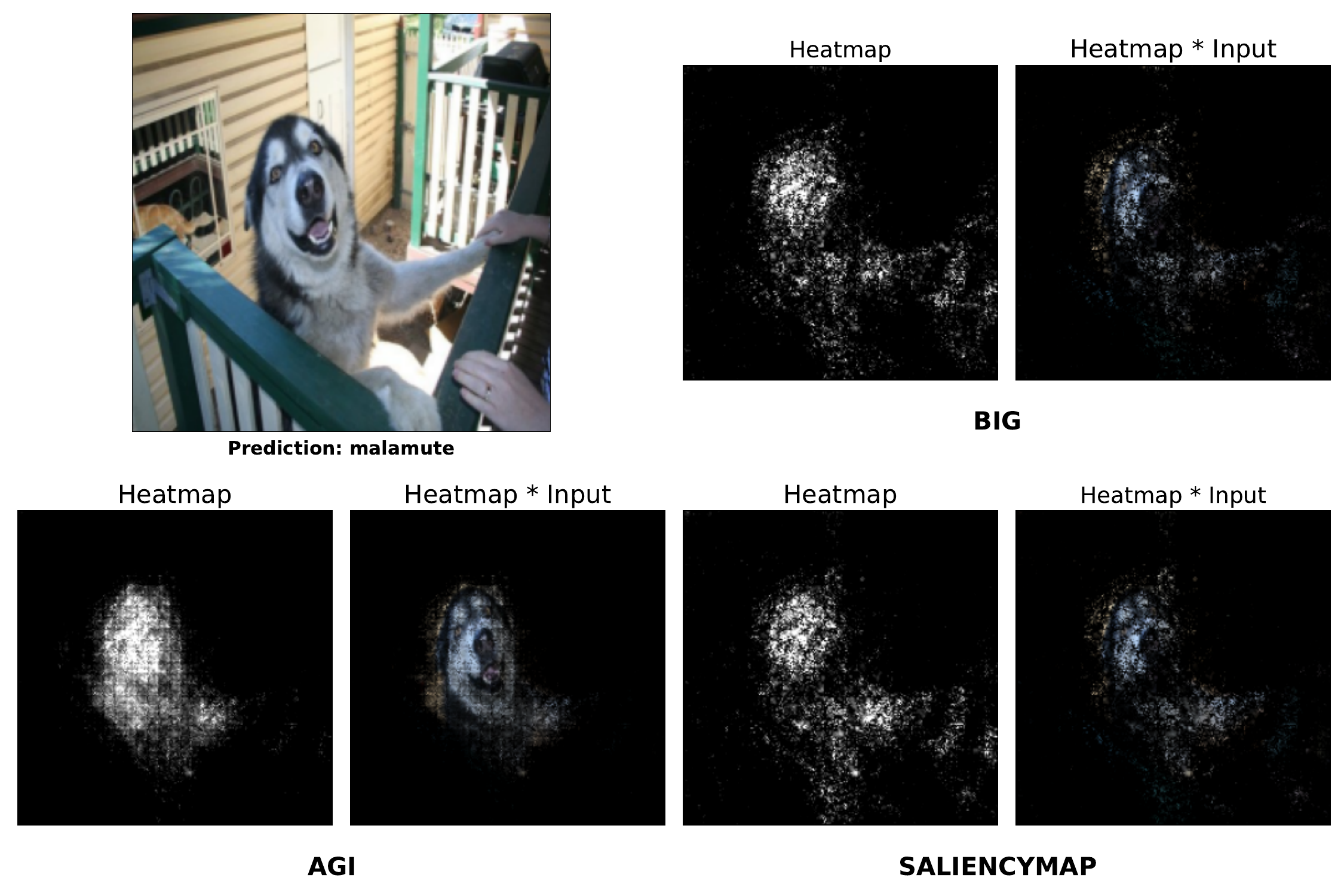}
}
\hfill
\subfigure[VGG-16 Interpretability Result]{
\includegraphics[width=0.48\textwidth]{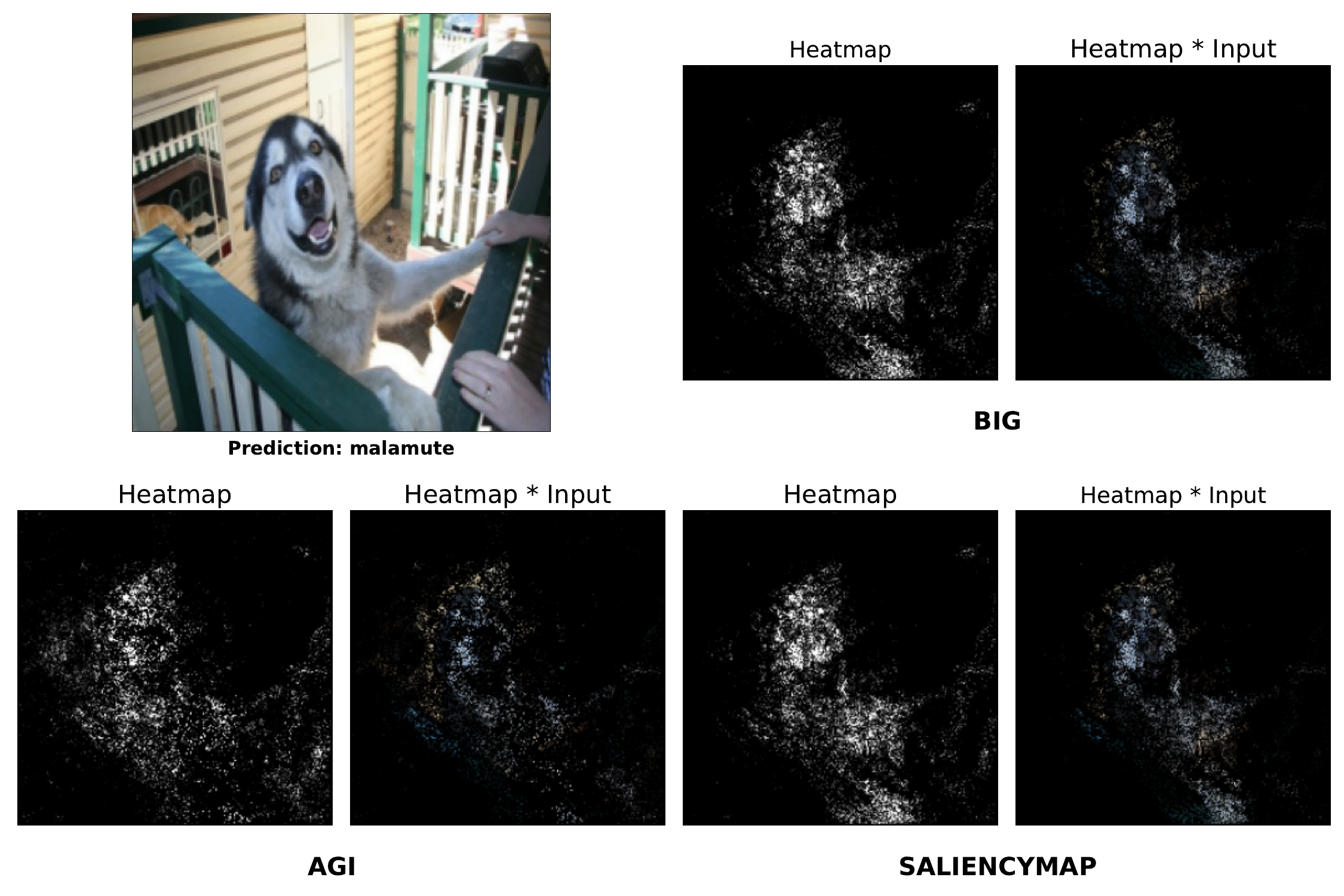}
}
\caption{Attribution Results of the Models}
\label{fig:module4}
\end{figure*}

\subsubsection{Interpretability analysis} In the experiments conducted in this module, we initially performed a global assessment of model interpretability using the Insertion Score and Deletion Score. As shown in Table~\ref{tab:module4}, we observed that ResNet-50 exhibits relatively higher Insertion Score and lower Deletion Score compared to VGG-16, indicating that ResNet-50 possesses better interpretability. Furthermore, based on Table~\ref{tab:module4}, we found that the AGI, BIG, and Saliency Map methods demonstrate relatively good attribution performance on both ResNet-50 and VGG-16. Therefore, we analyzed the heatmaps generated by these three methods for further analysis. As shown in Figure~\ref{fig:module4}, the white regions in the heatmaps represent the features that the model focuses on. From the top-left corner, which displays the original image, it can be observed that ResNet-50 exhibits a more concentrated focus on specific features compared to VGG-16, thus demonstrating better interpretability.

\begin{table*}[t]
\centering
\caption{Results of Utility evaluation of pruning models.}
\label{tab:module5}
\resizebox{\linewidth}{!}{%
\begin{tabular}{@{}c|c|c|ccccccccccccc@{}}
\toprule
Model & Method & Pruning Rate & Accuracy & Loss Value & TPR & TNR & PPV & NPV & FPR & FNR & FDR & ROC\_AUC & Precision & Recall & F1-Score \\ 
\midrule
\multirow{17}{*}{ResNet-50}
 & No pruning & 0 & 0.7929 & 3.8597 & 0.7929 & 0.9979 & 0.7936 & 0.9979 & 0.0021 & 0.2071 & 0.2064 & 0.9902 & 0.7936 & 0.7929 & 0.7922 \\
\cmidrule(l){2-16} 
 & \multirow{4}{*}{Taylor} & 0.35 & 0.7301 & 3.9557 & 0.7301 & 0.9973 & 0.7721 & 0.9973 & 0.0027 & 0.2699 & 0.2279 & 0.9877 & 0.7721 & 0.7301 & 0.7367 \\
 & & 0.4 & 0.6112 & 4.0869 & 0.6112 & 0.9961 & 0.817  & 0.9961 & 0.0039 & 0.3888 & 0.183 & 0.9798 & 0.817 & 0.6112 & 0.6602   \\
 & & 0.45 & 0.3058 & 4.3593 & 0.3058 & 0.993 & N/A & 0.993 & 0.007  & 0.6942 & N/A & 0.9313 & 0.8389 & 0.3058 & 0.3847 \\
 & & 0.5 & 0.0554 & 4.5603 & 0.0554 & 0.9905 & N/A & 0.9906 & 0.0095 & 0.9446 & N/A & 0.8338 & 0.3402 & 0.0554 & 0.0614 \\ 
\cmidrule(l){2-16} 
 & \multirow{4}{*}{ASL} & 0.35 & 0.6257 & 4.0277 & 0.6257 & 0.9962 & 0.7221 & 0.9962 & 0.0038 & 0.3743 & 0.2779 & 0.9717 & 0.7221 & 0.6257 & 0.6017 \\
 & & 0.4 & 0.5566 & 4.0994 & 0.5566 & 0.9955 & 0.7046 & 0.9955 & 0.0045 & 0.4434 & 0.2954 & 0.9608 & 0.7046 & 0.5566 & 0.5148 \\
 & & 0.45 & 0.5412 & 4.1178 & 0.5412 & 0.9954 & N/A & 0.9954 & 0.0046 & 0.4588 & N/A & 0.959 & 0.6716 & 0.5412 & 0.4821 \\
 & & 0.5 & 0.4944 & 4.1717 & 0.4944 & 0.9949 & N/A & 0.9949 & 0.0051 & 0.5056 & N/A & 0.953 & 0.5922 & 0.4944 & 0.4344 \\ 
\cmidrule(l){2-16} 
 & \multirow{4}{*}{OBD} & 0.35 & 0.7393 & 4.0965 & 0.7393 & 0.9974 & 0.7865 & 0.9974 & 0.0026 & 0.2607 & 0.2135 & 0.9898 & 0.7865 & 0.7393 & 0.7434 \\
 & & 0.4 & 0.6293 & 4.2932 & 0.6293 & 0.9963 & 0.8092 & 0.9963 & 0.0037 & 0.3707 & 0.1908 & 0.9886 & 0.8092 & 0.6293 & 0.6514 \\
 & & 0.45 & 0.4372 & 4.4456 & 0.4372 & 0.9943 & N/A & 0.9944 & 0.0057 & 0.5628 & N/A & 0.9873 & 0.8244 & 0.4372 & 0.4691 \\
 & & 0.5 & 0.2719 & 4.5201 & 0.2719 & 0.9926 & N/A & 0.9927 & 0.0074 & 0.7281 & N/A & 0.9851 & 0.6469 & 0.2719 & 0.281 \\ \cmidrule(l){2-16} 
 & \multirow{4}{*}{Greg-2} & 0.35 & 0.7879 & 3.866 & 0.7879 & 0.9979 & 0.7894 & 0.9979 & 0.0021 & 0.2121 & 0.2106 & 0.9903 & 0.7894 & 0.7879 & 0.7871 \\
 & & 0.4 & 0.7873 & 3.8704 & 0.7873 & 0.9979 & 0.7892 & 0.9979 & 0.0021 & 0.2127 & 0.2108 & 0.9903 & 0.7892 & 0.7873 & 0.7865 \\
 & & 0.45 & 0.7852 & 3.8768 & 0.7852 & 0.9978 & 0.7879 & 0.9978 & 0.0022 & 0.2148 & 0.2121 & 0.9902 & 0.7879 & 0.7852 & 0.7844 \\
 & & 0.5 & 0.7785 & 3.8882 & 0.7785 & 0.9978 & 0.7823 & 0.9978 & 0.0022 & 0.2215 & 0.2177 & 0.9901 & 0.7823 & 0.7785 & 0.7776 \\ \midrule
\multirow{17}{*}{VGG-16}
 & No pruning & 0 & 0.7259 & 3.91 & 0.7259 & 0.9972 & 0.7275 & 0.9972 & 0.0028 & 0.2741 & 0.2725 & 0.9855 & 0.7275 & 0.7259 & 0.7254 \\ 
\cmidrule(l){2-16} 
 & \multirow{4}{*}{Taylor} & 0.35 & 0.725 & 3.9109 & 0.725 & 0.9972 & 0.7266 & 0.9972 & 0.0028 & 0.275  & 0.2734 & 0.9852 & 0.7266 & 0.725 & 0.7247 \\
 & & 0.4 & 0.7243 & 3.9132 & 0.7243 & 0.9972 & 0.7259 & 0.9972 & 0.0028 & 0.2757 & 0.2741 & 0.9849 & 0.7259 & 0.7243 & 0.724 \\
 & & 0.45 & 0.7218 & 3.9187 & 0.7218 & 0.9972 & 0.7254 & 0.9972 & 0.0028 & 0.2782 & 0.2746 & 0.9843 & 0.7254 & 0.7218 & 0.7221 \\
 & & 0.5 & 0.7175 & 3.9317 & 0.7175 & 0.9971 & 0.726  & 0.9971 & 0.0029 & 0.2825 & 0.274  & 0.9829 & 0.726 & 0.7175 & 0.7186 \\ 
\cmidrule(l){2-16} 
 & \multirow{4}{*}{ASL} & 0.35 & 0.6585 & 3.9793 & 0.6585 & 0.9966 & 0.6773 & 0.9966 & 0.0034 & 0.3415 & 0.3227 & 0.9738 & 0.6773 & 0.6585 & 0.6562 \\
 & & 0.4 & 0.6255 & 4.0125 & 0.6255 & 0.9962 & 0.6555 & 0.9962 & 0.0038 & 0.3745 & 0.3445 & 0.9682 & 0.6555 & 0.6255 & 0.6225   \\
 & & 0.45 & 0.5786 & 4.0637 & 0.5786 & 0.9957 & 0.6319 & 0.9957 & 0.0043 & 0.4214 & 0.3681 & 0.9594 & 0.6319 & 0.5786 & 0.5766 \\
 & & 0.5 & 0.5435 & 4.1002 & 0.5435 & 0.9954 & 0.6168 & 0.9954 & 0.0046 & 0.4565 & 0.3832 & 0.9529 & 0.6168 & 0.5435 & 0.5413 \\ 
\cmidrule(l){2-16} 
 & \multirow{4}{*}{OBD} & 0.35 & 0.7262 & 3.9106 & 0.7262 & 0.9972 & 0.7277 & 0.9972 & 0.0028 & 0.2738 & 0.2723 & 0.9855 & 0.7277 & 0.7262 & 0.7256 \\
 & & 0.4 & 0.7259 & 3.9116 & 0.7259 & 0.9972 & 0.7275 & 0.9972 & 0.0028 & 0.2741 & 0.2725 & 0.9854 & 0.7275 & 0.7259 & 0.7253 \\
 & & 0.45 & 0.7264 & 3.9135 & 0.7264 & 0.9972 & 0.7281 & 0.9972 & 0.0028 & 0.2736 & 0.2719 & 0.9853 & 0.7281 & 0.7264 & 0.7259 \\
 & & 0.5 & 0.7261 & 3.917 & 0.7261 & 0.9972 & 0.7282 & 0.9972 & 0.0028 & 0.2739 & 0.2718 & 0.9852 & 0.7282 & 0.7261 & 0.7257 \\ \cmidrule(l){2-16} 
 & \multirow{4}{*}{Greg-2} & 0.35 & 0.723 & 3.912 & 0.723  & 0.9972 & 0.7248 & 0.9972 & 0.0028 & 0.277  & 0.2752 & 0.9859 & 0.7248 & 0.723  & 0.722 \\
 & & 0.4 & 0.721 & 3.9159 & 0.721  & 0.9972 & 0.7241 & 0.9972 & 0.0028 & 0.279  & 0.2759 & 0.9855 & 0.7241 & 0.721  & 0.7201 \\
 & & 0.45 & 0.7148 & 3.9204 & 0.7148 & 0.9971 & 0.7175 & 0.9971 & 0.0029 & 0.2852 & 0.2825 & 0.9851 & 0.7175 & 0.7148 & 0.7133 \\
 & & 0.5 & 0.7091 & 3.9281 & 0.7091 & 0.9971 & 0.7133 & 0.9971 & 0.0029 & 0.2909 & 0.2867 & 0.9844 & 0.7133 & 0.7091 & 0.7068 \\ 
\bottomrule
\end{tabular}%
}
\end{table*}

\subsubsection{Neuron analysis with pruning}
The purpose of this experiment is to evaluate the impact of different pruning rates on the performance of ResNet-50 and VGG-16 models. We employed four different pruning methods, including Taylor, ASL, OBD, and Greg-2, and observed the performance variations at different pruning rates. Firstly, we recorded the baseline performance of both models without pruning: ResNet-50 achieved an accuracy of 0.7929 and a loss value of 3.8597, while VGG-16 achieved an accuracy of 0.7259 and a loss value of 3.91.

Next, we conducted pruning experiments with different pruning rates and recorded the changes in performance metrics. In the ResNet-50 model, using the Taylor pruning method, when the pruning rate was set to 0.35, the accuracy dropped to 0.7301, and the loss value increased to 3.9557. As the pruning rate increased, the accuracy further declined, and the loss value continued to increase. When the pruning rate reached 0.5, the accuracy sharply dropped to 0.0554, and the loss value increased to 4.5603. In the VGG-16 model, using the Taylor pruning method, the accuracy gradually decreased and the loss value increased as the pruning rate increased. At a pruning rate of 0.5, the accuracy was 0.7175, and the loss value was 3.9317.

In addition to the Taylor pruning method, we also investigated the impact of ASL, OBD, and Greg-2 pruning methods on performance. For the ResNet-50 model, with ASL and OBD pruning methods, the accuracy gradually decreased as the pruning rate increased, while with the Greg-2 pruning method, the accuracy remained relatively stable, indicating a minor impact of pruning rate on performance. For the VGG-16 model, with ASL and OBD pruning methods, the accuracy gradually decreased as the pruning rate increased. While with the Greg-2 pruning method, similar to the ResNet-50 model, the accuracy remained relatively stable, indicating a minor impact of pruning rate on performance.


In summary, compared to VGG-16, ResNet-50 exhibits more pronounced performance fluctuations under the same pruning rate, indicating a larger proportion of critical parameters within the ResNet-50 model. This observation suggests that the neurons in ResNet-50 possess a lower degree of redundancy. As a deep residual network, ResNet-50 preserves more essential and refined data through skip connections and residual blocks, rendering it more susceptible to the effects of pruning methods. In contrast, VGG-16 adheres to a more traditional convolutional neural network architecture, potentially incorporating more redundant structures, which imparts greater tolerance to neuron pruning.

\subsubsection{Comprehensive Evaluation}

\begin{figure}[t]
\centering
\includegraphics[width=0.8\linewidth]{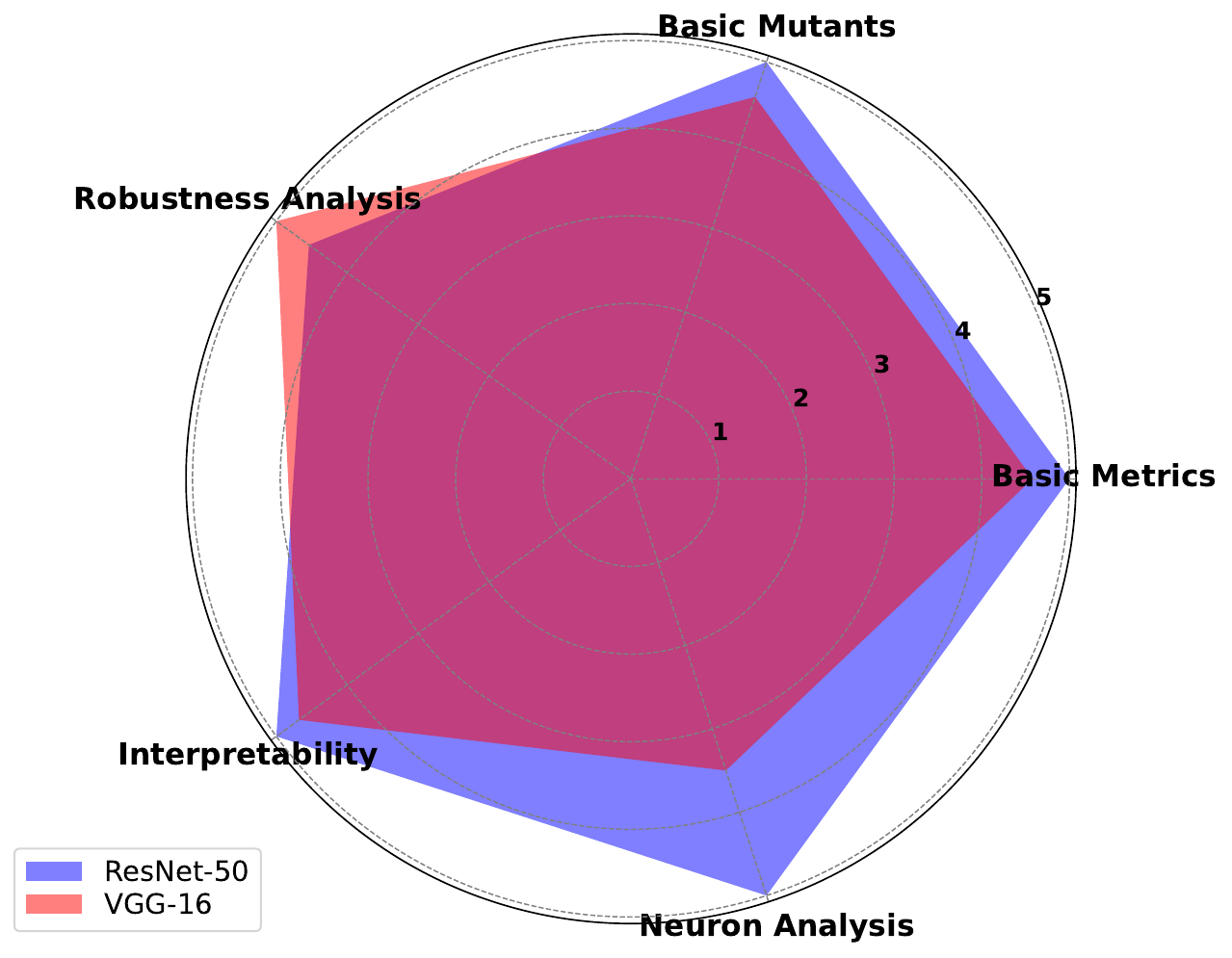}
\caption{Comprehensive Evaluation Result}
\label{fig:radar}
\end{figure}

Our approach allows for a comprehensive evaluation of models, enabling users to intuitively assess the performance of different competing models across five distinct modules. As depicted in Figure~\ref{fig:radar}, users can customize their selection of evaluation metrics for scoring sub-modules based on the emphasis of different tasks. In this particular example, we chose Precision as the evaluation metric in the Basic Metrics module, Precision under Label Error conditions in the Basic Mutants module, Average ASR of all attack methods in the Robustness Analysis module, Average Insertion Score in the Interpretability module, and in the Neural Analysis module, we select the maximum pruning rate that maintains model performance as our evaluation metric. We define the best model performance in the test set as a full score of 5 points, while other models are scored proportionally. Through this comprehensive assessment, it becomes evident that the ResNet-50 model outperforms the VGG-16 model across all aspects, except for Robustness.

\subsection{Answer to RQ2}
In this section, we demonstrate that the aforementioned modules support multimodal data, offering support for both text classification tasks and object detection tasks. We present the experimental results of \tool for these two tasks, showcasing its superior performance.

\subsubsection{Performance in text classification tasks}
As depicted in Table~\ref{tab:module11_text}, we present the evaluation of text classification models. The evaluation is conducted on TextCNN and AB-LSTM models using the SST-2 dataset. It can be observed that both models exhibit remarkably similar performance across most metrics. However, AB-LSTM outperforms TextCNN marginally in certain metrics such as Loss Value, Positive Predictive Value (PPV), Negative Predictive Value (NPV), False Positive Rate (FPR), False Negative Rate (FNR), and False Discovery Rate (FDR). Therefore, it can be inferred that the AB-LSTM model demonstrates slightly superior performance compared to the TextCNN model in this task.

\begin{table*}[t]
\centering
\caption{The results of basic metrics and mutant testing of the text classification}
\label{tab:module11_text}
\resizebox{\textwidth}{!}{%
\begin{tabular}{@{}c|c|c|c|ccccccccccccc@{}}
\toprule
Task                                 & Dataset                & Model                    & Method       & Accuracy & Loss Value & TPR    & TNR    & PPV    & NPV    & FPR    & FNR    & FDR    & ROC\_AUC & Precision & Recall & F1-Score \\ \midrule
\multirow{8}{*}{Text Classification} & \multirow{8}{*}{SST-2} & \multirow{4}{*}{TextCNN} & Origin Image & 0.84     & 0.4674     & 0.84   & 0.84   & 0.8401 & 0.8401 & 0.16   & 0.16   & 0.1599 & 0.9235   & 0.8401    & 0.84   & 0.84     \\
                                     &                        &                          & Label Error  & 0.7727   & 0.5299     & 0.7727 & 0.7727 & 0.7727 & 0.7727 & 0.2273 & 0.2273 & 0.2273 & 0.8383   & 0.7727    & 0.7727 & 0.7727   \\
                                     &                        &                          & Data Missing & 0.8222   & 0.4828     & 0.8222 & 0.8222 & 0.8223 & 0.8223 & 0.1778 & 0.1778 & 0.1777 & 0.9092   & 0.8223    & 0.8222 & 0.8222   \\
                                     &                        &                          & Data Shuffle & 0.84     & 0.4674     & 0.84   & 0.84   & 0.8401 & 0.8401 & 0.16   & 0.16   & 0.1599 & 0.9235   & 0.8401    & 0.84   & 0.84     \\ \cmidrule(l){3-17} 
                                     &                        & \multirow{4}{*}{AB-LSTM} & Origin Image & 0.8406   & 0.4645     & 0.8406 & 0.8406 & 0.8407 & 0.8407 & 0.1594 & 0.1594 & 0.1593 & 0.9212   & 0.8407    & 0.8406 & 0.8406   \\
                                     &                        &                          & Label Error  & 0.7664   & 0.5322     & 0.7666 & 0.7666 & 0.7665 & 0.7665 & 0.2334 & 0.2334 & 0.2335 & 0.8355   & 0.7665    & 0.7666 & 0.7664   \\
                                     &                        &                          & Data Missing & 0.8228   & 0.4798     & 0.8228 & 0.8228 & 0.8228 & 0.8228 & 0.1772 & 0.1772 & 0.1772 & 0.9053   & 0.8228    & 0.8228 & 0.8228   \\
                                     &                        &                          & Data Shuffle & 0.8406   & 0.4645     & 0.8406 & 0.8406 & 0.8407 & 0.8407 & 0.1594 & 0.1594 & 0.1593 & 0.9212   & 0.8407    & 0.8406 & 0.8406   \\ \bottomrule
\end{tabular}%
}
\end{table*}

As presented in Table~\ref{tab:module3_text}, the data generated by the Robustness module for testing text classification models is displayed. Based on the data in the table, we can observe that the AB-LSTM model demonstrates higher Attack Success Rates (ASR) under FGSM, SINI-FGSM, and PGD attacks. Conversely, the TextCNN model exhibits higher ASR under I-FGSM and MI-FGSM attacks. In general, the TextCNN model demonstrates a stable robustness, with a comparable defense performance across each attack method. On the other hand, the AB-LSTM model exhibits strong defense capabilities against certain attack methods while displaying weaker defense against others.

\begin{table}[t]
\centering
\caption{Table of model robustness evaluation results, the data in the table are ASR}
\label{tab:module3_text}
\resizebox{0.45\textwidth}{!}{%
\begin{tabular}{@{}l|l|l|ll@{}}
\toprule
Task                                  & Dataset                 & Model                    & Method    & ASR     \\ \midrule
\multirow{10}{*}{Text Classification} & \multirow{10}{*}{SST-2} & \multirow{5}{*}{TextCNN} & FGSM      & 86.60\% \\
                                      &                         &                          & I-FGSM    & 85.80\% \\
                                      &                         &                          & MI-FGSM   & 86.50\% \\
                                      &                         &                          & SINI-FGSM & 79.10\% \\
                                      &                         &                          & PGD       & 86.40\% \\ \cmidrule(l){3-5} 
                                      &                         & \multirow{5}{*}{AB-LSTM} & FGSM      & 97.70\% \\
                                      &                         &                          & I-FGSM    & 75.60\% \\
                                      &                         &                          & MI-FGSM   & 48.90\% \\
                                      &                         &                          & SINI-FGSM & 94.80\% \\
                                      &                         &                          & PGD       & 99.80\% \\ \bottomrule
\end{tabular}%
}
\end{table}

\begin{figure}
\centering
\includegraphics[width=\linewidth]{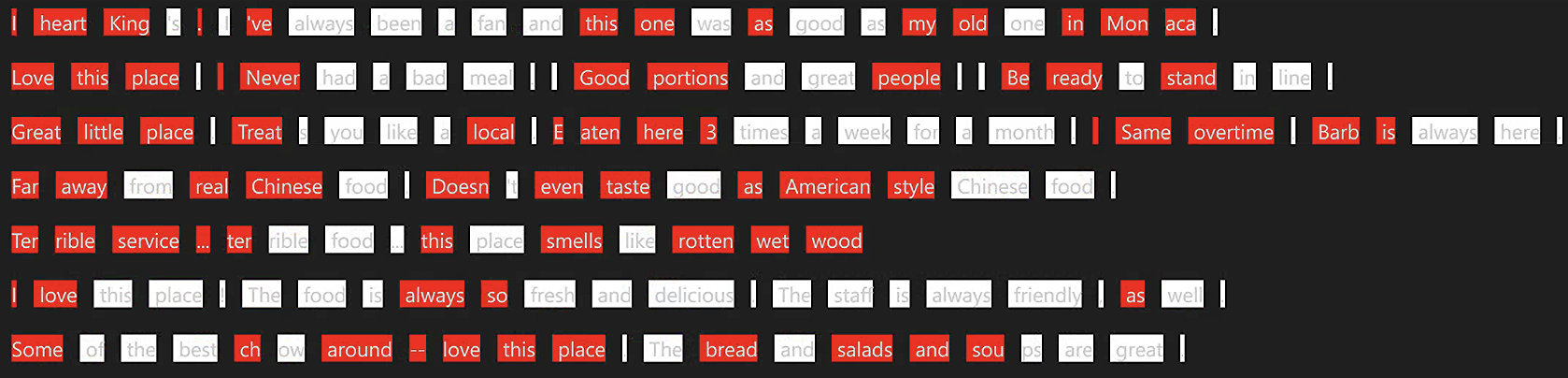}
\caption{Interpretability of the text classification model.}
\label{fig:interpretability_text}
\end{figure}

Figure~\ref{fig:interpretability_text} showcases the output results of the interpretability module for text classification tasks, wherein darker colors indicate a higher degree of model attention towards the corresponding regions of interest.

\subsubsection{Performance in object detection tasks}
Table~\ref{tab:module11_object} presents the evaluation of object detection tasks. In this section, the performance of Faster R-CNN and RetinaNet models was assessed using the COCO dataset. Both models exhibited relatively similar performance across different metrics. However, Faster R-CNN demonstrated a slight advantage in terms of overall average precision, while RetinaNet showcased a slight advantage in terms of overall average recall.

Table~\ref{tab:module3_object} presents the robustness evaluation of the object detection task. From the data, it can be observed that RetinaNet exhibits better robustness than Faster R-CNN, as it maintains higher average precision and average recall even after undergoing the same attacks. This indicates that RetinaNet demonstrates stronger resilience against attacks compared to Faster R-CNN.

\begin{table}[t]
\centering
\caption{The results of basic metrics and mutant testing of the object detection}
\label{tab:module11_object}
\resizebox{0.45\textwidth}{!}{%
\begin{tabular}{@{}c|c|c|ccc@{}}
\toprule
Task                               & Dataset                & Model                         & Method          & AP    & AR    \\ \midrule
\multirow{14}{*}{Object detection} & \multirow{14}{*}{COCO} & \multirow{7}{*}{Faster R-CNN} & Origin Image    & 0.585 & 0.508 \\
                                   &                        &                               & Data Missing    & 0.08  & 0.106 \\
                                   &                        &                               & Data Shuffle    & 0.585 & 0.508 \\
                                   &                        &                               & Noise Perturb   & 0.422 & 0.391 \\
                                   &                        &                               & Contrast Ratio  & 0.581 & 0.506 \\
                                   &                        &                               & Brightness      & 0.579 & 0.506 \\
                                   &                        &                               & Random Cropping & 0.548 & 0.433 \\ \cmidrule(l){3-6} 
                                   &                        & \multirow{7}{*}{RetinaNet}    & Origin image    & 0.557 & 0.537 \\
                                   &                        &                               & Data Missing    & 0.081 & 0.157 \\
                                   &                        &                               & Data Shuffle    & 0.557 & 0.537 \\
                                   &                        &                               & Noise Perturb   & 0.394 & 0.424 \\
                                   &                        &                               & Contrast Ratio  & 0.554 & 0.535 \\
                                   &                        &                               & Brightness      & 0.553 & 0.535 \\
                                   &                        &                               & Random Cropping & 0.523 & 0.461 \\ \bottomrule
\end{tabular}%
}
\end{table}

\begin{table*}[t]
\centering
\caption{The object detection model robustness evaluation results on COCO dataset.}
\label{tab:module3_object}
\resizebox{0.8\linewidth}{!}{%
\begin{tabular}{c|c|cc|cc|cc|cc}
\toprule
\multirow{2}{*}{Source Model}  & \multirow{2}{*}{Attack Method} & \multicolumn{2}{c|}{Faster R-CNN} & \multicolumn{2}{c|}{RetinaNet} & \multicolumn{2}{c|}{Mask R-CNN} & \multicolumn{2}{c}{SSD} \\ 
\cmidrule{3-10}
 & & AP & AR & AP & AR & AP & AR & AP & AR \\ 
\midrule
\multirow{10}{*}{Faster R-CNN} & Original & 0.585 & 0.508 & 0.557 & 0.537 & 0.591 & 0.519 & 0.415 & 0.365 \\
 & FGSM & 0.181 & 0.179 & 0.272 & 0.308 & 0.258 & 0.249 & 0.389 & 0.345 \\
 & I-FGSM & 0.051 & 0.075 & 0.164 & 0.239 & 0.134 & 0.164 & 0.394 & 0.351 \\
 & DI-FGSM & 0.109 & 0.146 & 0.211 & 0.276 & 0.189 & 0.218 & 0.38 & 0.34 \\
 & TI-FGSM & 0.161 & 0.187 & 0.321 & 0.381 & 0.296 & 0.31 & 0.376 & 0.336 \\
 & MI-FGSM & 0.041 & 0.066 & 0.119 & 0.178 & 0.098 & 0.12 & 0.375 & 0.335 \\
 & SINI-FGSM & 0.038 & 0.078 & 0.11 & 0.196 & 0.091 & 0.143 & 0.35 & 0.321 \\
 & PGD & 0.033 & 0.054 & 0.114 & 0.177 & 0.09 & 0.114 & 0.388 & 0.345 \\
 & SAA & 0.178 & 0.197 & 0.276 & 0.337 & 0.263 & 0.277 & 0.393 & 0.351 \\
 & Average & 0.153 & 0.166 & 0.238 & 0.292 & 0.223 & 0.235 & 0.384 & 0.343 \\ 
\midrule
\multirow{10}{*}{RetinaNet} & Original & 0.585 & 0.508 & 0.557 & 0.537 & 0.591 & 0.519 & 0.415 & 0.365 \\
 & FGSM & 0.296 & 0.292 & 0.169 & 0.206 & 0.306 & 0.305 & 0.393 & 0.347 \\
 & I-FGSM & 0.221 & 0.264 & 0.054 & 0.105 & 0.233 & 0.278 & 0.4 & 0.355 \\
 & DI-FGSM & 0.252 & 0.277 & 0.111 & 0.176 & 0.262 & 0.289 & 0.388 & 0.345 \\
 & TI-FGSM & 0.372 & 0.376 & 0.132 & 0.212 & 0.384 & 0.388 & 0.386 & 0.345 \\
 & MI-FGSM & 0.155 & 0.198 & 0.046 & 0.09 & 0.166 & 0.208 & 0.383 & 0.34 \\
 & SINI-FGSM & 0.145 & 0.2 & 0.047 & 0.098 & 0.157 & 0.211 & 0.359 & 0.326 \\
 & PGD & 0.166 & 0.212 & 0.04 & 0.08 & 0.176 & 0.22 & 0.395 & 0.35 \\
 & SAA & 0.305 & 0.319 & 0.162 & 0.233 & 0.322 & 0.335 & 0.396 & 0.353 \\
 & Average & 0.277 & 0.294 & 0.146 & 0.193 & 0.289 & 0.306 & 0.391 & 0.347 \\ 
\bottomrule
\end{tabular}%
}
\end{table*}

\subsection{Answer of RQ3}
By combining the Adversarial Robustness and Model Interpretability modules in \tool, we utilize pruning techniques to evaluate the performance of the model and provide optimization advice. Thus, the model complexity is reduced whilst the generalization ability is enhanced~\cite{li2022pls}. We have further investigated the impact of the pruning algorithm on model performance by combining the pruning results with the robustness and interpretability analysis.

\begin{figure}[t]
    \centering
    \includegraphics[width=\linewidth]{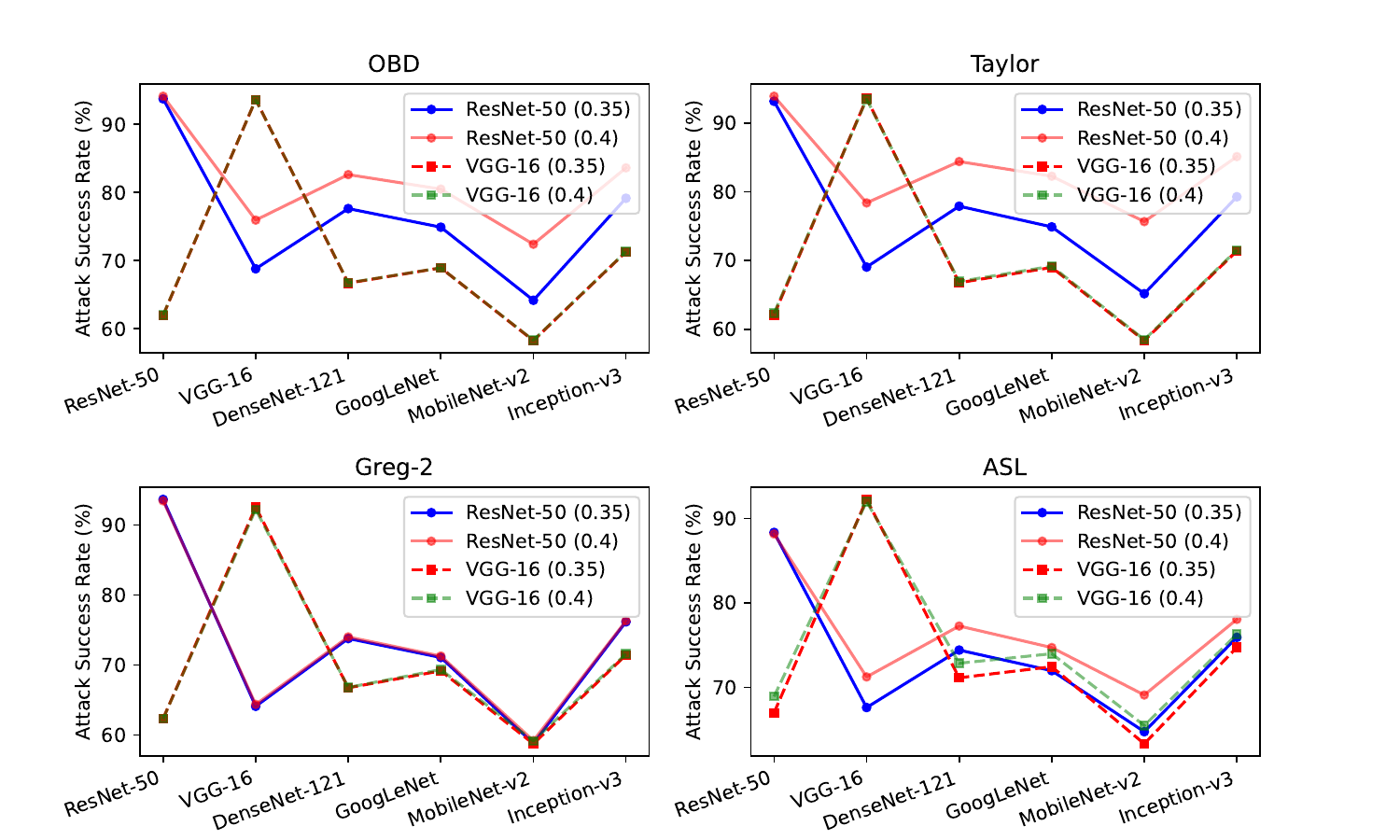}
    \caption{Average Attack Success Rate with different Pruning Rate}
    \label{fig:averageASR}
\end{figure}

\subsubsection{The impact of pruning on adversarial robustness}
As depicted in Figure~\ref{fig:averageASR}, we applied four different pruning algorithms to perform parameter pruning on ResNet-50 and VGG-16, achieving pruning ratios of 35\% and 40\% respectively. Observations indicate that, on the whole, VGG-16 exhibits superior robustness compared to ResNet-50. Following the pruning process, the robustness of VGG-16 remained almost unchanged, while ResNet-50 experienced a relatively substantial performance decline. Additionally, we noted that pruning the model using the Greg-2 method had minimal impact on the model's robustness.

\begin{figure}[t]
    \centering
    \includegraphics[width=\linewidth]{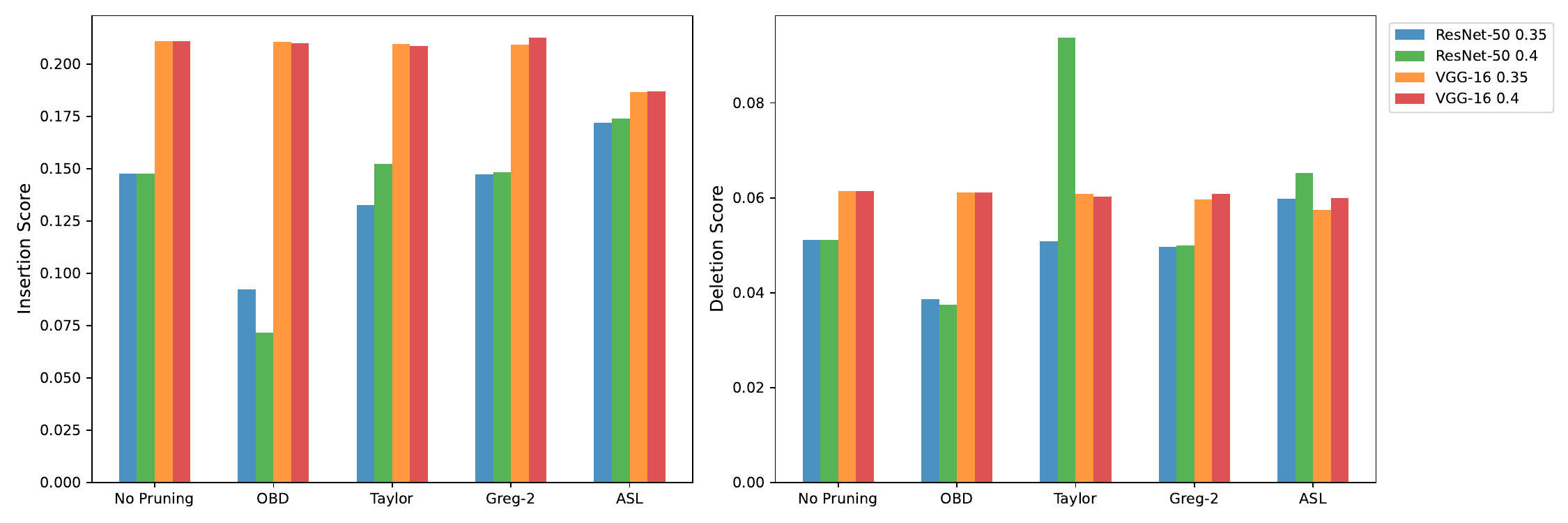}
    \caption{Average Insertion Score and Deletion Score with Different Pruning Rate. A higher Insertion Score indicates better model interpretability, while a lower Deletion Score indicates better interpretability. The comparison between Insertion Score and Deletion Score reflects the model's overall interpretability strength.}
    \label{fig:averageinsert}
\end{figure}
\subsubsection{The impact of pruning on model Interpretability}
As shown in Figure~\ref{fig:averageinsert}, we applied four different pruning algorithms to prune ResNet-50 and VGG-16 models by 35\% and 40\% respectively. It can be observed that compared to ResNet-50, VGG-16 exhibits a higher Insertion Score. However, the increase in Deletion Score for VGG-16 compared to ResNet-50 is only marginal. This suggests that in this experiment, VGG-16 demonstrates better interpretability than ResNet-50. It is worth noting that when using the Taylor method for pruning, ResNet-50 shows a sharp increase in Deletion Score, which could indicate that ResNet-50 is approaching its pruning limit. Additionally, when using the ASL pruning method, VGG-16 outperforms ResNet-50 in both Insertion Score and Deletion Score, indicating that VGG-16 exhibits superior interpretability across all aspects compared to ResNet-50 at this particular pruning stage.

\section{Conclusion}
In this paper, we proposed \tool, a comprehensive and effective multi-module testing tool for automated testing of DLS under the vast majority of testing requirements. In addition to the essential utility evaluation (including metric evaluation and mutation operations), \tool provides the measurements towards the adversarial robustness, model interpretability, and model’s neuron analysis to make an extensive report on the performance of DLS. Furthermore, the feasibility of \tool is tested in multi-modal scenarios. For tasks including image classification, text classification and object detection tasks, \tool shows superior performance and solves the ad-hoc problems of existing testing tools, indicating a high degree of scalability. Extensive experiments demonstrate that \tool is so far the state-of-the-art testing tool to build robust and trustworthy DLS.

\section*{Acknowledgement}
The work has been supported by the Cyber Security Research Centre Limited whose activities are partially funded by the Australian Government’s Cooperative Research Centres Programme.

\newpage
\bibliographystyle{IEEEtran}
\bibliography{main}

\begin{thebibliography}{100}
\providecommand{\url}[1]{#1}
\csname url@samestyle\endcsname
\providecommand{\newblock}{\relax}
\providecommand{\bibinfo}[2]{#2}
\providecommand{\BIBentrySTDinterwordspacing}{\spaceskip=0pt\relax}
\providecommand{\BIBentryALTinterwordstretchfactor}{4}
\providecommand{\BIBentryALTinterwordspacing}{\spaceskip=\fontdimen2\font plus
\BIBentryALTinterwordstretchfactor\fontdimen3\font minus
  \fontdimen4\font\relax}
\providecommand{\BIBforeignlanguage}[2]{{%
\expandafter\ifx\csname l@#1\endcsname\relax
\typeout{** WARNING: IEEEtran.bst: No hyphenation pattern has been}%
\typeout{** loaded for the language `#1'. Using the pattern for}%
\typeout{** the default language instead.}%
\else
\language=\csname l@#1\endcsname
\fi
#2}}
\providecommand{\BIBdecl}{\relax}
\BIBdecl

\bibitem{he2016deep}
K.~He, X.~Zhang, S.~Ren, and J.~Sun, ``Deep residual learning for image
  recognition,'' in \emph{Proceedings of the IEEE conference on computer vision
  and pattern recognition}, 2016, pp. 770--778.

\bibitem{simonyan2014very}
K.~Simonyan and A.~Zisserman, ``Very deep convolutional networks for
  large-scale image recognition,'' \emph{arXiv preprint arXiv:1409.1556}, 2014.

\bibitem{xiong2016achieving}
W.~Xiong, J.~Droppo, X.~Huang, F.~Seide, M.~Seltzer, A.~Stolcke, D.~Yu, and
  G.~Zweig, ``Achieving human parity in conversational speech recognition,''
  \emph{arXiv preprint arXiv:1610.05256}, 2016.

\bibitem{chen2020software}
J.~Chen, K.~Hu, Y.~Yu, Z.~Chen, Q.~Xuan, Y.~Liu, and V.~Filkov, ``Software
  visualization and deep transfer learning for effective software defect
  prediction,'' in \emph{Proceedings of the ACM/IEEE 42nd international
  conference on software engineering}, 2020, pp. 578--589.

\bibitem{devanbu2020deep}
P.~Devanbu, M.~Dwyer, S.~Elbaum, M.~Lowry, K.~Moran, D.~Poshyvanyk, B.~Ray,
  R.~Singh, and X.~Zhang, ``Deep learning \& software engineering: State of
  research and future directions,'' \emph{arXiv preprint arXiv:2009.08525},
  2020.

\bibitem{li2021deeppayload}
Y.~Li, J.~Hua, H.~Wang, C.~Chen, and Y.~Liu, ``Deeppayload: Black-box backdoor
  attack on deep learning models through neural payload injection,'' in
  \emph{2021 IEEE/ACM 43rd International Conference on Software Engineering
  (ICSE)}.\hskip 1em plus 0.5em minus 0.4em\relax IEEE, 2021, pp. 263--274.

\bibitem{mirabella2021deep}
A.~G. Mirabella, A.~Martin-Lopez, S.~Segura, L.~Valencia-Cabrera, and
  A.~Ruiz-Cort{\'e}s, ``Deep learning-based prediction of test input validity
  for restful apis,'' in \emph{2021 IEEE/ACM Third International Workshop on
  Deep Learning for Testing and Testing for Deep Learning (DeepTest)}.\hskip
  1em plus 0.5em minus 0.4em\relax IEEE, 2021, pp. 9--16.

\bibitem{sedaghatbaf2021automated}
A.~Sedaghatbaf, M.~H. Moghadam, and M.~Saadatmand, ``Automated performance
  testing based on active deep learning,'' in \emph{2021 IEEE/ACM International
  Conference on Automation of Software Test (AST)}.\hskip 1em plus 0.5em minus
  0.4em\relax IEEE, 2021, pp. 11--19.

\bibitem{xiao2021development}
B.~Xiao and S.-C. Kang, ``Development of an image data set of construction
  machines for deep learning object detection,'' \emph{Journal of Computing in
  Civil Engineering}, vol.~35, no.~2, p. 05020005, 2021.

\bibitem{shrestha2019review}
A.~Shrestha and A.~Mahmood, ``Review of deep learning algorithms and
  architectures,'' \emph{IEEE access}, vol.~7, pp. 53\,040--53\,065, 2019.

\bibitem{parvat2017survey}
A.~Parvat, J.~Chavan, S.~Kadam, S.~Dev, and V.~Pathak, ``A survey of
  deep-learning frameworks,'' in \emph{2017 International Conference on
  Inventive Systems and Control (ICISC)}.\hskip 1em plus 0.5em minus
  0.4em\relax IEEE, 2017, pp. 1--7.

\bibitem{dhar2023challenges}
T.~Dhar, N.~Dey, S.~Borra, and R.~S. Sherratt, ``Challenges of deep learning in
  medical image analysis—improving explainability and trust,'' \emph{IEEE
  Transactions on Technology and Society}, vol.~4, no.~1, pp. 68--75, 2023.

\bibitem{kavitha2023ant}
R.~Kavitha, D.~K. Jothi, K.~Saravanan, M.~P. Swain, J.~L.~A. Gonz{\'a}les,
  R.~J. Bhardwaj, E.~Adomako \emph{et~al.}, ``Ant colony optimization-enabled
  cnn deep learning technique for accurate detection of cervical cancer,''
  \emph{BioMed Research International}, vol. 2023, 2023.

\bibitem{lee2023end}
D.-H. Lee and J.-L. Liu, ``End-to-end deep learning of lane detection and path
  prediction for real-time autonomous driving,'' \emph{Signal, Image and Video
  Processing}, vol.~17, no.~1, pp. 199--205, 2023.

\bibitem{alaba2023deep}
S.~Y. Alaba and J.~E. Ball, ``Deep learning-based image 3-d object detection
  for autonomous driving,'' \emph{IEEE Sensors Journal}, vol.~23, no.~4, pp.
  3378--3394, 2023.

\bibitem{haluza2023artificial}
D.~Haluza and D.~Jungwirth, ``Artificial intelligence and ten societal
  megatrends: An exploratory study using gpt-3,'' \emph{Systems}, vol.~11,
  no.~3, p. 120, 2023.

\bibitem{maddigan2023chat2vis}
P.~Maddigan and T.~Susnjak, ``Chat2vis: Generating data visualisations via
  natural language using chatgpt, codex and gpt-3 large language models,''
  \emph{IEEE Access}, 2023.

\bibitem{zohdinasab2023efficient}
T.~Zohdinasab, V.~Riccio, A.~Gambi, and P.~Tonella, ``Efficient and effective
  feature space exploration for testing deep learning systems,'' \emph{ACM
  Transactions on Software Engineering and Methodology}, vol.~32, no.~2, pp.
  1--38, 2023.

\bibitem{sekhon2019towards}
J.~Sekhon and C.~Fleming, ``Towards improved testing for deep learning,'' in
  \emph{2019 IEEE/ACM 41st International Conference on Software Engineering:
  New Ideas and Emerging Results (ICSE-NIER)}.\hskip 1em plus 0.5em minus
  0.4em\relax IEEE, 2019, pp. 85--88.

\bibitem{gerasimou2020importance}
S.~Gerasimou, H.~F. Eniser, A.~Sen, and A.~Cakan, ``Importance-driven deep
  learning system testing,'' in \emph{Proceedings of the ACM/IEEE 42nd
  International Conference on Software Engineering}, 2020, pp. 702--713.

\bibitem{wadawadagi2020sentiment}
R.~Wadawadagi and V.~Pagi, ``Sentiment analysis with deep neural networks:
  comparative study and performance assessment,'' \emph{Artificial Intelligence
  Review}, vol.~53, no.~8, pp. 6155--6195, 2020.

\bibitem{jin2023novel}
Y.~Jin, Z.~Li, C.~Qin, J.~Liu, Y.~Liu, L.~Zhao, and C.~Liu, ``A novel
  attentional deep neural network-based assessment method for ecg quality,''
  \emph{Biomedical Signal Processing and Control}, vol.~79, p. 104064, 2023.

\bibitem{alwadi2023framework}
M.~Alwadi, G.~Chetty, and M.~Yamin, ``A framework for vehicle quality
  evaluation based on interpretable machine learning,'' \emph{International
  Journal of Information Technology}, vol.~15, no.~1, pp. 129--136, 2023.

\bibitem{guo2023comprehensive}
J.~Guo, W.~Bao, J.~Wang, Y.~Ma, X.~Gao, G.~Xiao, A.~Liu, J.~Dong, X.~Liu, and
  W.~Wu, ``A comprehensive evaluation framework for deep model robustness,''
  \emph{Pattern Recognition}, vol. 137, p. 109308, 2023.

\bibitem{gajera2023comprehensive}
H.~K. Gajera, D.~R. Nayak, and M.~A. Zaveri, ``A comprehensive analysis of
  dermoscopy images for melanoma detection via deep cnn features,''
  \emph{Biomedical Signal Processing and Control}, vol.~79, p. 104186, 2023.

\bibitem{sawhney2023comparative}
R.~Sawhney, A.~Malik, S.~Sharma, and V.~Narayan, ``A comparative assessment of
  artificial intelligence models used for early prediction and evaluation of
  chronic kidney disease,'' \emph{Decision Analytics Journal}, vol.~6, p.
  100169, 2023.

\bibitem{finlayson2018adversarial}
S.~G. Finlayson, H.~W. Chung, I.~S. Kohane, and A.~L. Beam, ``Adversarial
  attacks against medical deep learning systems,'' \emph{arXiv preprint
  arXiv:1804.05296}, 2018.

\bibitem{nesti2022evaluating}
F.~Nesti, G.~Rossolini, S.~Nair, A.~Biondi, and G.~Buttazzo, ``Evaluating the
  robustness of semantic segmentation for autonomous driving against real-world
  adversarial patch attacks,'' in \emph{Proceedings of the IEEE/CVF Winter
  Conference on Applications of Computer Vision}, 2022, pp. 2280--2289.

\bibitem{gu2017badnets}
T.~Gu, B.~Dolan-Gavitt, and S.~Garg, ``Badnets: Identifying vulnerabilities in
  the machine learning model supply chain,'' \emph{arXiv preprint
  arXiv:1708.06733}, 2017.

\bibitem{mehtab2022analysis}
S.~Mehtab and J.~Sen, ``Analysis and forecasting of financial time series using
  cnn and lstm-based deep learning models,'' in \emph{Advances in Distributed
  Computing and Machine Learning: Proceedings of ICADCML 2021}.\hskip 1em plus
  0.5em minus 0.4em\relax Springer, 2022, pp. 405--423.

\bibitem{qiu2022multimodal}
S.~Qiu, M.~I. Miller, P.~S. Joshi, J.~C. Lee, C.~Xue, Y.~Ni, Y.~Wang,
  I.~De~Anda-Duran, P.~H. Hwang, J.~A. Cramer \emph{et~al.}, ``Multimodal deep
  learning for alzheimer’s disease dementia assessment,'' \emph{Nature
  communications}, vol.~13, no.~1, p. 3404, 2022.

\bibitem{zhang2022adversarial}
Q.~Zhang, S.~Hu, J.~Sun, Q.~A. Chen, and Z.~M. Mao, ``On adversarial robustness
  of trajectory prediction for autonomous vehicles,'' in \emph{Proceedings of
  the IEEE/CVF Conference on Computer Vision and Pattern Recognition}, 2022,
  pp. 15\,159--15\,168.

\bibitem{dieber2022novel}
J.~Dieber and S.~Kirrane, ``A novel model usability evaluation framework (muse)
  for explainable artificial intelligence,'' \emph{Information Fusion},
  vol.~81, pp. 143--153, 2022.

\bibitem{sajjad2022neuron}
H.~Sajjad, N.~Durrani, and F.~Dalvi, ``Neuron-level interpretation of deep nlp
  models: A survey,'' \emph{Transactions of the Association for Computational
  Linguistics}, vol.~10, pp. 1285--1303, 2022.

\bibitem{zhou2022understanding}
D.~Zhou, Z.~Yu, E.~Xie, C.~Xiao, A.~Anandkumar, J.~Feng, and J.~M. Alvarez,
  ``Understanding the robustness in vision transformers,'' in
  \emph{International Conference on Machine Learning}.\hskip 1em plus 0.5em
  minus 0.4em\relax PMLR, 2022, pp. 27\,378--27\,394.

\bibitem{dai2022comprehensive}
E.~Dai, T.~Zhao, H.~Zhu, J.~Xu, Z.~Guo, H.~Liu, J.~Tang, and S.~Wang, ``A
  comprehensive survey on trustworthy graph neural networks: Privacy,
  robustness, fairness, and explainability,'' \emph{arXiv preprint
  arXiv:2204.08570}, 2022.

\bibitem{guan2022explaining}
Y.~Guan, F.~Lecue, J.~Chen, R.~Li, and J.~Z. Pan, ``Explaining image
  classification through knowledge-aware neuron interpretation,'' 2022.

\bibitem{yang2022revisiting}
Z.~Yang, J.~Shi, M.~H. Asyrofi, and D.~Lo, ``Revisiting neuron coverage metrics
  and quality of deep neural networks,'' in \emph{2022 IEEE International
  Conference on Software Analysis, Evolution and Reengineering (SANER)}.\hskip
  1em plus 0.5em minus 0.4em\relax IEEE, 2022, pp. 408--419.

\bibitem{pawlowski2017dltk}
N.~Pawlowski, S.~I. Ktena, M.~C. Lee, B.~Kainz, D.~Rueckert, B.~Glocker, and
  M.~Rajchl, ``Dltk: State of the art reference implementations for deep
  learning on medical images,'' \emph{arXiv preprint arXiv:1711.06853}, 2017.

\bibitem{pei2017deepxplore}
K.~Pei, Y.~Cao, J.~Yang, and S.~Jana, ``Deepxplore: Automated whitebox testing
  of deep learning systems,'' in \emph{proceedings of the 26th Symposium on
  Operating Systems Principles}, 2017, pp. 1--18.

\bibitem{tian2018deeptest}
Y.~Tian, K.~Pei, S.~Jana, and B.~Ray, ``Deeptest: Automated testing of
  deep-neural-network-driven autonomous cars,'' in \emph{Proceedings of the
  40th international conference on software engineering}, 2018, pp. 303--314.

\bibitem{weber2023less}
D.~Weber, F.~Merkle, P.~Sch{\"o}ttle, and S.~Schl{\"o}gl, ``Less is more: The
  influence of pruning on the explainability of cnns,'' \emph{arXiv preprint
  arXiv:2302.08878}, 2023.

\bibitem{jordao2021effect}
A.~Jordao and H.~Pedrini, ``On the effect of pruning on adversarial
  robustness,'' in \emph{Proceedings of the IEEE/CVF International Conference
  on Computer Vision}, 2021, pp. 1--11.

\bibitem{ma2018deepmutation}
L.~Ma, F.~Zhang, J.~Sun, M.~Xue, B.~Li, F.~Juefei-Xu, C.~Xie, L.~Li, Y.~Liu,
  J.~Zhao \emph{et~al.}, ``Deepmutation: Mutation testing of deep learning
  systems,'' in \emph{2018 IEEE 29th international symposium on software
  reliability engineering (ISSRE)}.\hskip 1em plus 0.5em minus 0.4em\relax
  IEEE, 2018, pp. 100--111.

\bibitem{hu2019deepmutation++}
Q.~Hu, L.~Ma, X.~Xie, B.~Yu, Y.~Liu, and J.~Zhao, ``Deepmutation++: A mutation
  testing framework for deep learning systems,'' in \emph{2019 34th IEEE/ACM
  International Conference on Automated Software Engineering (ASE)}.\hskip 1em
  plus 0.5em minus 0.4em\relax IEEE, 2019, pp. 1158--1161.

\bibitem{jin2023poster}
Z.~Jin, Z.~Zhu, H.~Hu, M.~Xue, and H.~Chen, ``Poster: Ml-compass: A
  comprehensive assessment framework for machine learning models,'' in
  \emph{Proceedings of the 2023 ACM Asia Conference on Computer and
  Communications Security}, 2023, pp. 1031--1033.

\bibitem{zheng2022neuronfair}
H.~Zheng, Z.~Chen, T.~Du, X.~Zhang, Y.~Cheng, S.~Ji, J.~Wang, Y.~Yu, and
  J.~Chen, ``Neuronfair: Interpretable white-box fairness testing through
  biased neuron identification,'' in \emph{Proceedings of the 44th
  International Conference on Software Engineering}, 2022, pp. 1519--1531.

\bibitem{ma2018deepgauge}
L.~Ma, F.~Juefei-Xu, F.~Zhang, J.~Sun, M.~Xue, B.~Li, C.~Chen, T.~Su, L.~Li,
  Y.~Liu \emph{et~al.}, ``Deepgauge: Multi-granularity testing criteria for
  deep learning systems,'' in \emph{Proceedings of the 33rd ACM/IEEE
  international conference on automated software engineering}, 2018, pp.
  120--131.

\bibitem{li2022interpretdl}
X.~Li, H.~Xiong, X.~Li, X.~Wu, Z.~Chen, and D.~Dou, ``Interpretdl: Explaining
  deep models in paddlepaddle,'' \emph{Journal of Machine Learning Research},
  vol.~23, no. 197, pp. 1--6, 2022.

\bibitem{nicolae2018adversarial}
M.-I. Nicolae, M.~Sinn, M.~N. Tran, B.~Buesser, A.~Rawat, M.~Wistuba,
  V.~Zantedeschi, N.~Baracaldo, B.~Chen, H.~Ludwig \emph{et~al.}, ``Adversarial
  robustness toolbox v1. 0.0,'' \emph{arXiv preprint arXiv:1807.01069}, 2018.

\bibitem{goodfellow2014explaining}
I.~J. Goodfellow, J.~Shlens, and C.~Szegedy, ``Explaining and harnessing
  adversarial examples,'' \emph{arXiv preprint arXiv:1412.6572}, 2014.

\bibitem{moosavi2016deepfool}
S.-M. Moosavi-Dezfooli, A.~Fawzi, and P.~Frossard, ``Deepfool: a simple and
  accurate method to fool deep neural networks,'' in \emph{Proceedings of the
  IEEE conference on computer vision and pattern recognition}, 2016, pp.
  2574--2582.

\bibitem{carlini2017towards}
N.~Carlini and D.~Wagner, ``Towards evaluating the robustness of neural
  networks,'' in \emph{2017 ieee symposium on security and privacy (sp)}.\hskip
  1em plus 0.5em minus 0.4em\relax Ieee, 2017, pp. 39--57.

\bibitem{morris2020textattack}
J.~X. Morris, E.~Lifland, J.~Y. Yoo, J.~Grigsby, D.~Jin, and Y.~Qi,
  ``Textattack: A framework for adversarial attacks, data augmentation, and
  adversarial training in nlp,'' \emph{arXiv preprint arXiv:2005.05909}, 2020.

\bibitem{ebrahimi2017hotflip}
J.~Ebrahimi, A.~Rao, D.~Lowd, and D.~Dou, ``Hotflip: White-box adversarial
  examples for text classification,'' \emph{arXiv preprint arXiv:1712.06751},
  2017.

\bibitem{jin2020bert}
D.~Jin, Z.~Jin, J.~T. Zhou, and P.~Szolovits, ``Is bert really robust? a strong
  baseline for natural language attack on text classification and entailment,''
  in \emph{Proceedings of the AAAI conference on artificial intelligence},
  vol.~34, no.~05, 2020, pp. 8018--8025.

\bibitem{rauber2017foolbox}
J.~Rauber, W.~Brendel, and M.~Bethge, ``Foolbox: A python toolbox to benchmark
  the robustness of machine learning models,'' \emph{arXiv preprint
  arXiv:1707.04131}, 2017.

\bibitem{li2018textbugger}
J.~Li, S.~Ji, T.~Du, B.~Li, and T.~Wang, ``Textbugger: Generating adversarial
  text against real-world applications,'' \emph{arXiv preprint
  arXiv:1812.05271}, 2018.

\bibitem{dong2019there}
Y.~Dong, P.~Zhang, J.~Wang, S.~Liu, J.~Sun, J.~Hao, X.~Wang, L.~Wang, J.~S.
  Dong, and D.~Ting, ``There is limited correlation between coverage and
  robustness for deep neural networks,'' \emph{arXiv preprint
  arXiv:1911.05904}, 2019.

\bibitem{harel2020neuron}
F.~Harel-Canada, L.~Wang, M.~A. Gulzar, Q.~Gu, and M.~Kim, ``Is neuron coverage
  a meaningful measure for testing deep neural networks?'' in \emph{Proceedings
  of the 28th ACM Joint Meeting on European Software Engineering Conference and
  Symposium on the Foundations of Software Engineering}, 2020, pp. 851--862.

\bibitem{jin2023excitement}
H.~Jin, R.~Chen, H.~Zheng, J.~Chen, Y.~Cheng, Y.~Yu, T.~Chen, and X.~Liu,
  ``Excitement surfeited turns to errors: Deep learning testing framework based
  on excitable neurons,'' \emph{Information Sciences}, vol. 637, p. 118936,
  2023.

\bibitem{li2020learning}
X.~Li, X.~Li, D.~Pan, and D.~Zhu, ``On the learning property of logistic and
  softmax losses for deep neural networks,'' in \emph{Proceedings of the AAAI
  Conference on Artificial Intelligence}, vol.~34, no.~04, 2020, pp.
  4739--4746.

\bibitem{seltzer2013investigation}
M.~L. Seltzer, D.~Yu, and Y.~Wang, ``An investigation of deep neural networks
  for noise robust speech recognition,'' in \emph{2013 IEEE international
  conference on acoustics, speech and signal processing}.\hskip 1em plus 0.5em
  minus 0.4em\relax IEEE, 2013, pp. 7398--7402.

\bibitem{yin2017comparative}
W.~Yin, K.~Kann, M.~Yu, and H.~Sch{\"u}tze, ``Comparative study of cnn and rnn
  for natural language processing,'' \emph{arXiv preprint arXiv:1702.01923},
  2017.

\bibitem{lian2018xdeepfm}
J.~Lian, X.~Zhou, F.~Zhang, Z.~Chen, X.~Xie, and G.~Sun, ``xdeepfm: Combining
  explicit and implicit feature interactions for recommender systems,'' in
  \emph{Proceedings of the 24th ACM SIGKDD international conference on
  knowledge discovery \& data mining}, 2018, pp. 1754--1763.

\bibitem{jindal2016learning}
I.~Jindal, M.~Nokleby, and X.~Chen, ``Learning deep networks from noisy labels
  with dropout regularization,'' in \emph{2016 IEEE 16th International
  Conference on Data Mining (ICDM)}.\hskip 1em plus 0.5em minus 0.4em\relax
  IEEE, 2016, pp. 967--972.

\bibitem{dixon2017classification}
M.~Dixon, D.~Klabjan, and J.~H. Bang, ``Classification-based financial markets
  prediction using deep neural networks,'' \emph{Algorithmic Finance}, vol.~6,
  no. 3-4, pp. 67--77, 2017.

\bibitem{guan2020analysis}
S.~Guan and M.~Loew, ``Analysis of generalizability of deep neural networks
  based on the complexity of decision boundary,'' in \emph{2020 19th IEEE
  International Conference on Machine Learning and Applications (ICMLA)}.\hskip
  1em plus 0.5em minus 0.4em\relax IEEE, 2020, pp. 101--106.

\bibitem{biggio2013evasion}
B.~Biggio, I.~Corona, D.~Maiorca, B.~Nelson, N.~{\v{S}}rndi{\'c}, P.~Laskov,
  G.~Giacinto, and F.~Roli, ``Evasion attacks against machine learning at test
  time,'' in \emph{Machine Learning and Knowledge Discovery in Databases:
  European Conference, ECML PKDD 2013, Prague, Czech Republic, September 23-27,
  2013, Proceedings, Part III 13}.\hskip 1em plus 0.5em minus 0.4em\relax
  Springer, 2013, pp. 387--402.

\bibitem{su2019one}
J.~Su, D.~V. Vargas, and K.~Sakurai, ``One pixel attack for fooling deep neural
  networks,'' \emph{IEEE Transactions on Evolutionary Computation}, vol.~23,
  no.~5, pp. 828--841, 2019.

\bibitem{deng2020analysis}
Y.~Deng, X.~Zheng, T.~Zhang, C.~Chen, G.~Lou, and M.~Kim, ``An analysis of
  adversarial attacks and defenses on autonomous driving models,'' in
  \emph{2020 IEEE international conference on pervasive computing and
  communications (PerCom)}.\hskip 1em plus 0.5em minus 0.4em\relax IEEE, 2020,
  pp. 1--10.

\bibitem{huq2020analysis}
A.~Huq and M.~T. Pervin, ``Analysis of adversarial attacks on skin cancer
  recognition,'' in \emph{2020 International Conference on Data Science and Its
  Applications (ICoDSA)}.\hskip 1em plus 0.5em minus 0.4em\relax IEEE, 2020,
  pp. 1--4.

\bibitem{kurakin2018adversarial}
A.~Kurakin, I.~J. Goodfellow, and S.~Bengio, ``Adversarial examples in the
  physical world,'' in \emph{Artificial intelligence safety and
  security}.\hskip 1em plus 0.5em minus 0.4em\relax Chapman and Hall/CRC, 2018,
  pp. 99--112.

\bibitem{madry2017towards}
A.~Madry, A.~Makelov, L.~Schmidt, D.~Tsipras, and A.~Vladu, ``Towards deep
  learning models resistant to adversarial attacks,'' \emph{arXiv preprint
  arXiv:1706.06083}, 2017.

\bibitem{dong2017discovering}
Y.~Dong, F.~Liao, T.~Pang, X.~Hu, and J.~Zhu, ``Discovering adversarial
  examples with momentum,'' \emph{arXiv preprint arXiv:1710.06081}, vol.~5,
  2017.

\bibitem{dong2019evading}
Y.~Dong, T.~Pang, H.~Su, and J.~Zhu, ``Evading defenses to transferable
  adversarial examples by translation-invariant attacks,'' in \emph{Proceedings
  of the IEEE/CVF Conference on Computer Vision and Pattern Recognition}, 2019,
  pp. 4312--4321.

\bibitem{xiao2018generating}
C.~Xiao, B.~Li, J.-Y. Zhu, W.~He, M.~Liu, and D.~Song, ``Generating adversarial
  examples with adversarial networks,'' \emph{arXiv preprint arXiv:1801.02610},
  2018.

\bibitem{li2020qeba}
H.~Li, X.~Xu, X.~Zhang, S.~Yang, and B.~Li, ``Qeba: Query-efficient
  boundary-based blackbox attack,'' in \emph{Proceedings of the IEEE/CVF
  conference on computer vision and pattern recognition}, 2020, pp. 1221--1230.

\bibitem{chen2017zoo}
P.-Y. Chen, H.~Zhang, Y.~Sharma, J.~Yi, and C.-J. Hsieh, ``Zoo: Zeroth order
  optimization based black-box attacks to deep neural networks without training
  substitute models,'' in \emph{Proceedings of the 10th ACM workshop on
  artificial intelligence and security}, 2017, pp. 15--26.

\bibitem{lin2019nesterov}
J.~Lin, C.~Song, K.~He, L.~Wang, and J.~E. Hopcroft, ``Nesterov accelerated
  gradient and scale invariance for adversarial attacks,'' \emph{arXiv preprint
  arXiv:1908.06281}, 2019.

\bibitem{xie2019improving}
C.~Xie, Z.~Zhang, Y.~Zhou, S.~Bai, J.~Wang, Z.~Ren, and A.~L. Yuille,
  ``Improving transferability of adversarial examples with input diversity,''
  in \emph{Proceedings of the IEEE/CVF Conference on Computer Vision and
  Pattern Recognition}, 2019, pp. 2730--2739.

\bibitem{gao2020patch}
L.~Gao, Q.~Zhang, J.~Song, X.~Liu, and H.~T. Shen, ``Patch-wise attack for
  fooling deep neural network,'' in \emph{Computer Vision--ECCV 2020: 16th
  European Conference, Glasgow, UK, August 23--28, 2020, Proceedings, Part
  XXVIII 16}.\hskip 1em plus 0.5em minus 0.4em\relax Springer, 2020, pp.
  307--322.

\bibitem{zhang2022improving}
J.~Zhang, W.~Wu, J.-t. Huang, Y.~Huang, W.~Wang, Y.~Su, and M.~R. Lyu,
  ``Improving adversarial transferability via neuron attribution-based
  attacks,'' in \emph{Proceedings of the IEEE/CVF Conference on Computer Vision
  and Pattern Recognition}, 2022, pp. 14\,993--15\,002.

\bibitem{long2022frequency}
Y.~Long, Q.~Zhang, B.~Zeng, L.~Gao, X.~Liu, J.~Zhang, and J.~Song, ``Frequency
  domain model augmentation for adversarial attack,'' in \emph{Computer
  Vision--ECCV 2022: 17th European Conference, Tel Aviv, Israel, October
  23--27, 2022, Proceedings, Part IV}.\hskip 1em plus 0.5em minus 0.4em\relax
  Springer, 2022, pp. 549--566.

\bibitem{jin2023danaa}
Z.~Jin, Z.~Zhu, X.~Wang, J.~Zhang, J.~Shen, and H.~Chen, ``Danaa: Towards
  transferable attacks with double adversarial neuron attribution,''
  \emph{arXiv preprint arXiv:2310.10427}, 2023.

\bibitem{10.1145/3583780.3614927}
\BIBentryALTinterwordspacing
Z.~Zhu, H.~Chen, J.~Zhang, X.~Wang, Z.~Jin, Q.~Lu, J.~Shen, and K.-K.~R. Choo,
  ``Improving adversarial transferability via frequency-based stationary point
  search,'' in \emph{Proceedings of the 32nd ACM International Conference on
  Information and Knowledge Management}, ser. CIKM '23.\hskip 1em plus 0.5em
  minus 0.4em\relax New York, NY, USA: Association for Computing Machinery,
  2023, p. 3626–3635. [Online]. Available:
  \url{https://doi.org/10.1145/3583780.3614927}
\BIBentrySTDinterwordspacing

\bibitem{qin2022boosting}
Z.~Qin, Y.~Fan, Y.~Liu, L.~Shen, Y.~Zhang, J.~Wang, and B.~Wu, ``Boosting the
  transferability of adversarial attacks with reverse adversarial
  perturbation,'' \emph{arXiv preprint arXiv:2210.05968}, 2022.

\bibitem{adadi2018peeking}
A.~Adadi and M.~Berrada, ``Peeking inside the black-box: a survey on
  explainable artificial intelligence (xai),'' \emph{IEEE access}, vol.~6, pp.
  52\,138--52\,160, 2018.

\bibitem{ribeiro2016should}
M.~T. Ribeiro, S.~Singh, and C.~Guestrin, ``" why should i trust you?"
  explaining the predictions of any classifier,'' in \emph{Proceedings of the
  22nd ACM SIGKDD international conference on knowledge discovery and data
  mining}, 2016, pp. 1135--1144.

\bibitem{shrikumar2017learning}
A.~Shrikumar, P.~Greenside, and A.~Kundaje, ``Learning important features
  through propagating activation differences,'' in \emph{International
  conference on machine learning}.\hskip 1em plus 0.5em minus 0.4em\relax PMLR,
  2017, pp. 3145--3153.

\bibitem{lundberg2017unified}
S.~M. Lundberg and S.-I. Lee, ``A unified approach to interpreting model
  predictions,'' \emph{Advances in neural information processing systems},
  vol.~30, 2017.

\bibitem{fong2017interpretable}
R.~C. Fong and A.~Vedaldi, ``Interpretable explanations of black boxes by
  meaningful perturbation,'' in \emph{Proceedings of the IEEE international
  conference on computer vision}, 2017, pp. 3429--3437.

\bibitem{datta2016algorithmic}
A.~Datta, S.~Sen, and Y.~Zick, ``Algorithmic transparency via quantitative
  input influence: Theory and experiments with learning systems,'' in
  \emph{2016 IEEE symposium on security and privacy (SP)}.\hskip 1em plus 0.5em
  minus 0.4em\relax IEEE, 2016, pp. 598--617.

\bibitem{li2016understanding}
J.~Li, W.~Monroe, and D.~Jurafsky, ``Understanding neural networks through
  representation erasure,'' \emph{arXiv preprint arXiv:1612.08220}, 2016.

\bibitem{selvaraju2017grad}
R.~R. Selvaraju, M.~Cogswell, A.~Das, R.~Vedantam, D.~Parikh, and D.~Batra,
  ``Grad-cam: Visual explanations from deep networks via gradient-based
  localization,'' in \emph{Proceedings of the IEEE international conference on
  computer vision}, 2017, pp. 618--626.

\bibitem{wang2020score}
H.~Wang, Z.~Wang, M.~Du, F.~Yang, Z.~Zhang, S.~Ding, P.~Mardziel, and X.~Hu,
  ``Score-cam: Score-weighted visual explanations for convolutional neural
  networks,'' in \emph{Proceedings of the IEEE/CVF conference on computer
  vision and pattern recognition workshops}, 2020, pp. 24--25.

\bibitem{patra2020incremental}
A.~Patra and J.~A. Noble, ``Incremental learning of fetal heart anatomies using
  interpretable saliency maps,'' in \emph{Annual Conference on Medical Image
  Understanding and Analysis}.\hskip 1em plus 0.5em minus 0.4em\relax Springer,
  2020, pp. 129--141.

\bibitem{springenberg2014striving}
J.~T. Springenberg, A.~Dosovitskiy, T.~Brox, and M.~Riedmiller, ``Striving for
  simplicity: The all convolutional net,'' \emph{arXiv preprint
  arXiv:1412.6806}, 2014.

\bibitem{sundararajan2017axiomatic}
M.~Sundararajan, A.~Taly, and Q.~Yan, ``Axiomatic attribution for deep
  networks,'' in \emph{International conference on machine learning}.\hskip 1em
  plus 0.5em minus 0.4em\relax PMLR, 2017, pp. 3319--3328.

\bibitem{erion2021improving}
G.~Erion, J.~D. Janizek, P.~Sturmfels, S.~M. Lundberg, and S.-I. Lee,
  ``Improving performance of deep learning models with axiomatic attribution
  priors and expected gradients,'' \emph{Nature machine intelligence}, vol.~3,
  no.~7, pp. 620--631, 2021.

\bibitem{wang2021robust}
Z.~Wang, M.~Fredrikson, and A.~Datta, ``Robust models are more interpretable
  because attributions look normal,'' \emph{arXiv preprint arXiv:2103.11257},
  2021.

\bibitem{pan2021explaining}
D.~Pan, X.~Li, and D.~Zhu, ``Explaining deep neural network models with
  adversarial gradient integration,'' in \emph{Thirtieth International Joint
  Conference on Artificial Intelligence (IJCAI)}, 2021.

\bibitem{10.1145/3583780.3614889}
\BIBentryALTinterwordspacing
Z.~Zhu, H.~Chen, Z.~Jin, X.~Wang, J.~Zhang, M.~Xue, Q.~Lu, J.~Shen, and
  K.-K.~R. Choo, ``Fvw: Finding valuable weight on deep neural network for
  model pruning,'' in \emph{Proceedings of the 32nd ACM International
  Conference on Information and Knowledge Management}, ser. CIKM '23.\hskip 1em
  plus 0.5em minus 0.4em\relax New York, NY, USA: Association for Computing
  Machinery, 2023, p. 3657–3666. [Online]. Available:
  \url{https://doi.org/10.1145/3583780.3614889}
\BIBentrySTDinterwordspacing

\bibitem{ye2019adversarial}
S.~Ye, K.~Xu, S.~Liu, H.~Cheng, J.-H. Lambrechts, H.~Zhang, A.~Zhou, K.~Ma,
  Y.~Wang, and X.~Lin, ``Adversarial robustness vs. model compression, or
  both?'' in \emph{Proceedings of the IEEE/CVF International Conference on
  Computer Vision}, 2019, pp. 111--120.

\bibitem{liu2018fine}
K.~Liu, B.~Dolan-Gavitt, and S.~Garg, ``Fine-pruning: Defending against
  backdooring attacks on deep neural networks,'' in \emph{Research in Attacks,
  Intrusions, and Defenses: 21st International Symposium, RAID 2018, Heraklion,
  Crete, Greece, September 10-12, 2018, Proceedings 21}.\hskip 1em plus 0.5em
  minus 0.4em\relax Springer, 2018, pp. 273--294.

\bibitem{xu2021efficient}
Z.~Xu, J.~Sun, Y.~Liu, and G.~Sun, ``An efficient channel-level pruning for
  cnns without fine-tuning,'' in \emph{2021 International Joint Conference on
  Neural Networks (IJCNN)}.\hskip 1em plus 0.5em minus 0.4em\relax IEEE, 2021,
  pp. 1--8.

\bibitem{hu2016network}
H.~Hu, R.~Peng, Y.-W. Tai, and C.-K. Tang, ``Network trimming: A data-driven
  neuron pruning approach towards efficient deep architectures,'' \emph{arXiv
  preprint arXiv:1607.03250}, 2016.

\bibitem{lecun1989optimal}
Y.~LeCun, J.~Denker, and S.~Solla, ``Optimal brain damage,'' \emph{Advances in
  neural information processing systems}, vol.~2, 1989.

\bibitem{molchanov2019importance}
P.~Molchanov, A.~Mallya, S.~Tyree, I.~Frosio, and J.~Kautz, ``Importance
  estimation for neural network pruning,'' in \emph{Proceedings of the IEEE/CVF
  conference on computer vision and pattern recognition}, 2019, pp.
  11\,264--11\,272.

\bibitem{wang2020neural}
H.~Wang, C.~Qin, Y.~Zhang, and Y.~Fu, ``Neural pruning via growing
  regularization,'' \emph{arXiv preprint arXiv:2012.09243}, 2020.

\bibitem{retsinas2020weight}
G.~Retsinas, A.~Elafrou, G.~Goumas, and P.~Maragos, ``Weight pruning via
  adaptive sparsity loss,'' \emph{arXiv preprint arXiv:2006.02768}, 2020.

\bibitem{kapishnikov2021guided}
A.~Kapishnikov, S.~Venugopalan, B.~Avci, B.~Wedin, M.~Terry, and T.~Bolukbasi,
  ``Guided integrated gradients: An adaptive path method for removing noise,''
  in \emph{Proceedings of the IEEE/CVF conference on computer vision and
  pattern recognition}, 2021, pp. 5050--5058.

\bibitem{petsiuk2018rise}
V.~Petsiuk, A.~Das, and K.~Saenko, ``Rise: Randomized input sampling for
  explanation of black-box models,'' \emph{arXiv preprint arXiv:1806.07421},
  2018.

\bibitem{krizhevsky2009learning}
A.~Krizhevsky, G.~Hinton \emph{et~al.}, ``Learning multiple layers of features
  from tiny images,'' 2009.

\bibitem{lin2014microsoft}
T.-Y. Lin, M.~Maire, S.~Belongie, J.~Hays, P.~Perona, D.~Ramanan,
  P.~Doll{\'a}r, and C.~L. Zitnick, ``Microsoft coco: Common objects in
  context,'' in \emph{Computer Vision--ECCV 2014: 13th European Conference,
  Zurich, Switzerland, September 6-12, 2014, Proceedings, Part V 13}.\hskip 1em
  plus 0.5em minus 0.4em\relax Springer, 2014, pp. 740--755.

\bibitem{socher2013recursive}
R.~Socher, A.~Perelygin, J.~Wu, J.~Chuang, C.~D. Manning, A.~Y. Ng, and
  C.~Potts, ``Recursive deep models for semantic compositionality over a
  sentiment treebank,'' in \emph{Proceedings of the 2013 conference on
  empirical methods in natural language processing}, 2013, pp. 1631--1642.

\bibitem{kim2014convolutional}
Y.~Kim, ``Convolutional neural networks for sentence classification,''
  \emph{arXiv preprint arXiv:1408.5882}, 2014.

\bibitem{liu2019ab}
Z.~Liu, W.~Zhou, and H.~Li, ``Ab-lstm: Attention-based bidirectional lstm model
  for scene text detection,'' \emph{ACM Transactions on Multimedia Computing,
  Communications, and Applications (TOMM)}, vol.~15, no.~4, pp. 1--23, 2019.

\bibitem{ren2015faster}
S.~Ren, K.~He, R.~Girshick, and J.~Sun, ``Faster r-cnn: Towards real-time
  object detection with region proposal networks,'' \emph{Advances in neural
  information processing systems}, vol.~28, 2015.

\bibitem{lin2017focal}
T.-Y. Lin, P.~Goyal, R.~Girshick, K.~He, and P.~Doll{\'a}r, ``Focal loss for
  dense object detection,'' in \emph{Proceedings of the IEEE international
  conference on computer vision}, 2017, pp. 2980--2988.

\bibitem{li2022pls}
W.~Li, X.~Wang, H.~Han, and J.~Qiao, ``A pls-based pruning algorithm for
  simplified long--short term memory neural network in time series
  prediction,'' \emph{Knowledge-Based Systems}, vol. 254, p. 109608, 2022.

\end{thebibliography}

\newpage
\begin{IEEEbiography}
[{\includegraphics[width=1in,height=1.25in,keepaspectratio]{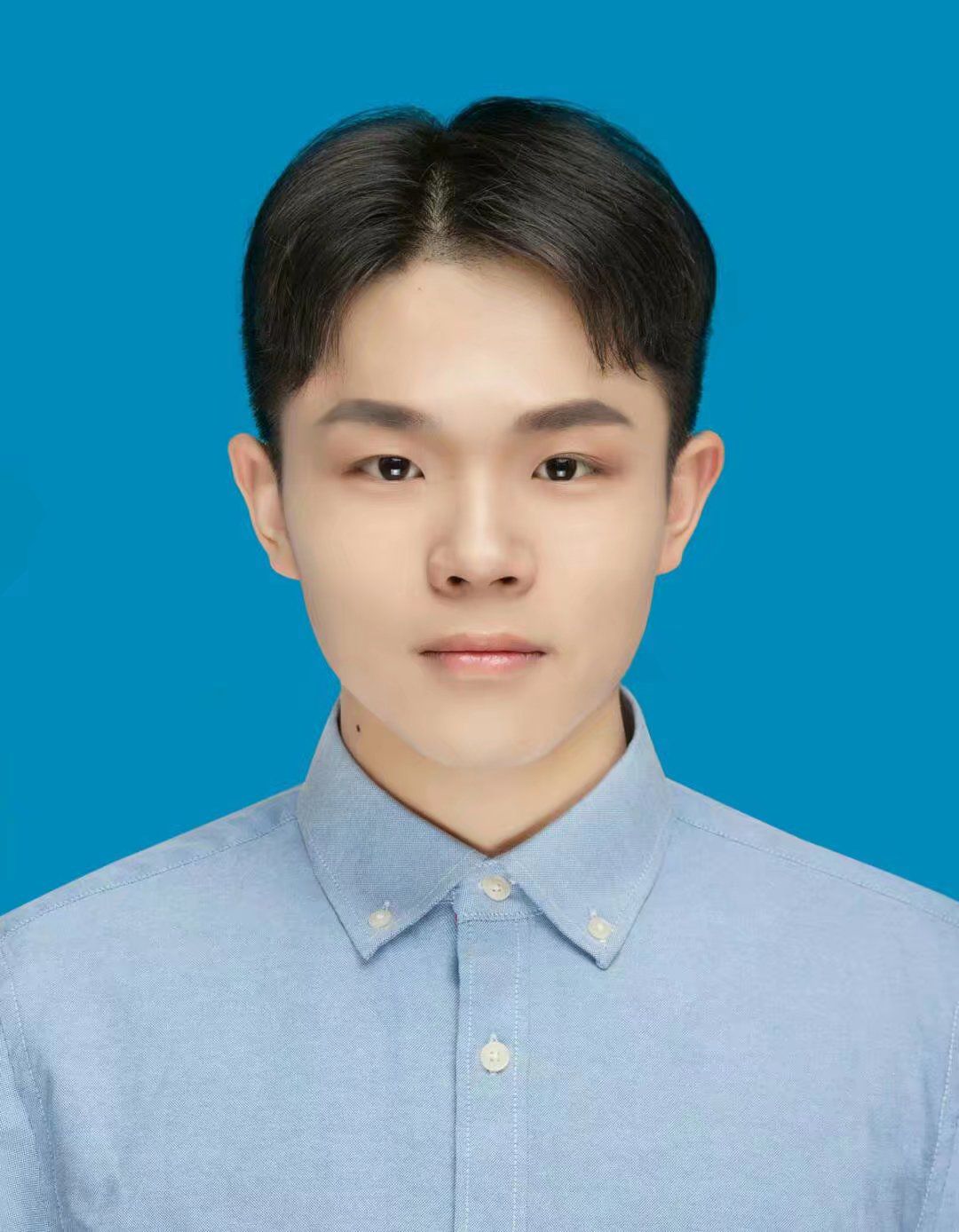}}]{Zhiyu Zhu}
 is currently pursuing the Master degree at the University of Sydney, he is also an Honours Student at CSIRO Data61. His research interests encompass AI model security and Trustworthy Machine Learning.
\end{IEEEbiography}

\begin{IEEEbiography}
[{\includegraphics[width=1in,height=1.25in,keepaspectratio]{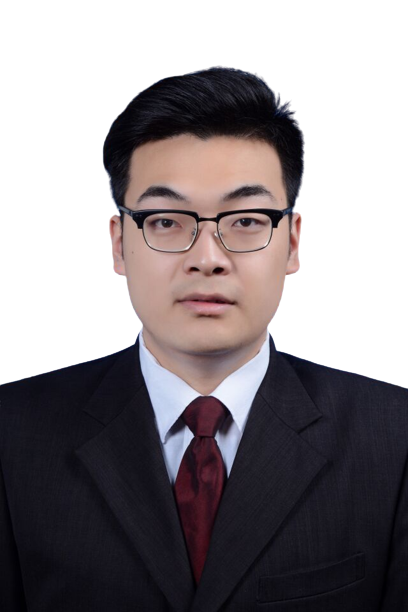}}]{Zhibo Jin}
received his bachelor's degree from China University of Petroleum (Beijing) and is currently pursuing a Master's degree at the University of Sydney. He is also an Honours Student at CSIRO Data61. His research focuses on AI model security and Trustworthy Machine Learning.
\end{IEEEbiography}

\begin{IEEEbiography}
[{\includegraphics[width=1in,height=1.25in,keepaspectratio]{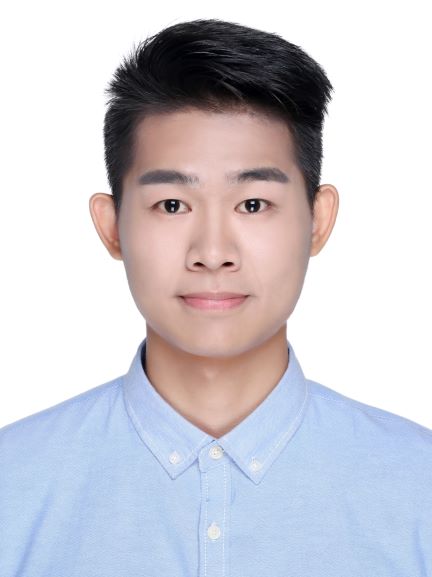}}]{Hongsheng Hu}
Hongsheng Hu received his PhD degree at Faculty of Engineering, University of Auckland, New Zealand. His research focuses on AI privacy and security, especially membership inference attacks, differential privacy, and inference attacks in the context of federated learning. He has published 8 international refereed journal and conference papers, including ACM Computing Surveys, IJCAI, and ICDM.
\end{IEEEbiography}

\begin{IEEEbiography}
[{\includegraphics[width=1in,height=1.25in,keepaspectratio]{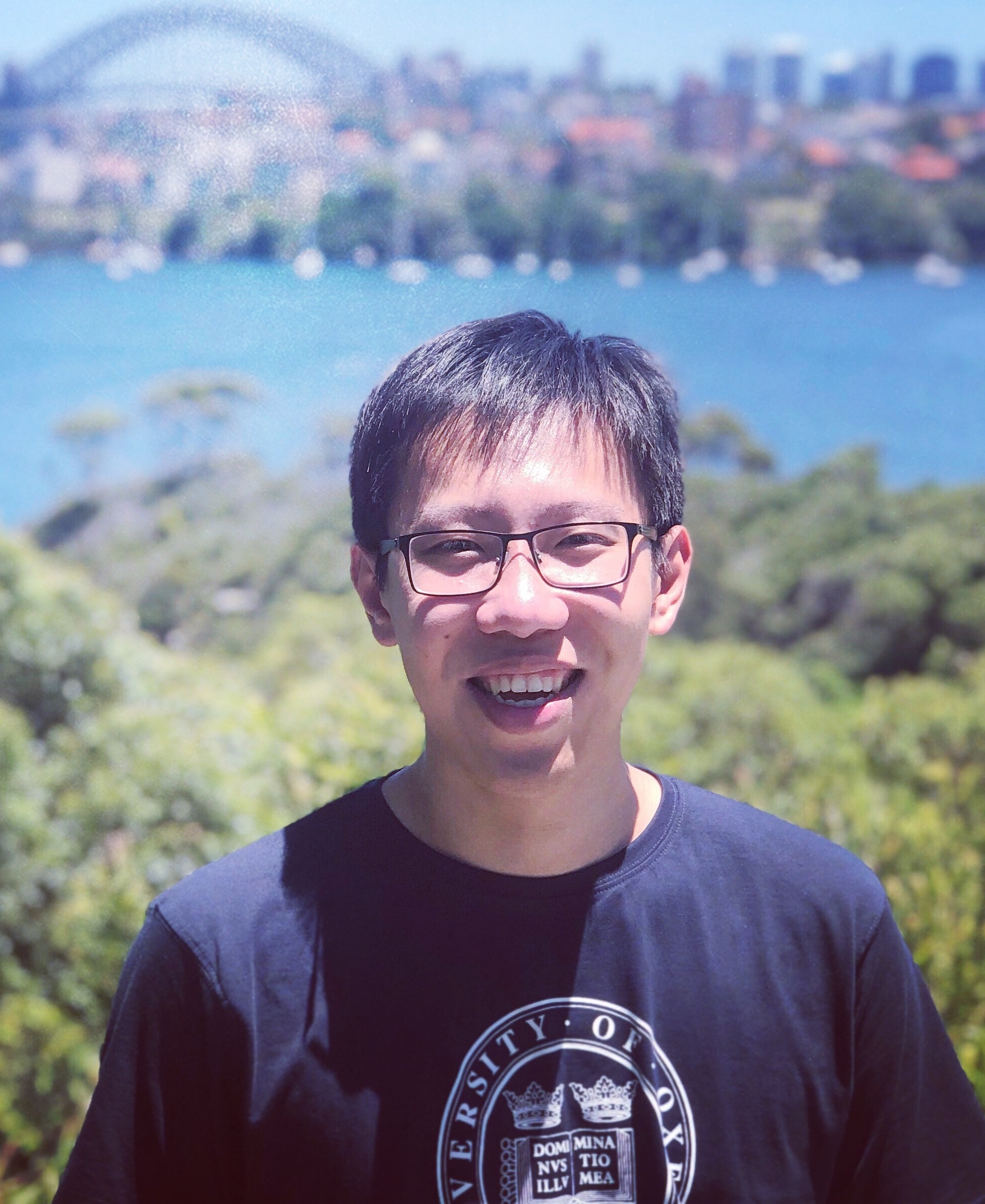}}]{Minhui Xue}
received the Ph.D. degree from East China Normal University, Shanghai, China. He is currently a Senior Research Scientist with CSIRO’s Data61, Marsfield, NSW, Australia. He is also an Honorary Lecturer with Macquarie University, Sydney, NSW, Australia. His current research interests include machine learning security and privacy, system security and privacy, and Internet measurement. Dr. Xue was a recipient of the ACM Special Interest Group on Software Engineering (ACM CCS) Best Paper Award Runner-Up and the ACM Special Interest Group on Software Engineering (SIGSOFT) Distinguished Paper Award. He serves or has served on the Program Committees of the 2021 and 2023 IEEE Symposium on Security and Privacy; the 2022 and 2023 ACM Conference on Computer and Communications Security; the 2023 and 2024 USENIX Security Symposium (USENIX) Security Symposium; the 2023 and 2024 Network and Distributed System Security Symposium; and the 2021, 2022, and 2023 ACM/IEEE International Conference on Software Engineering.
\end{IEEEbiography}

\begin{IEEEbiography}
[{\includegraphics[width=1in,height=1.25in,keepaspectratio]{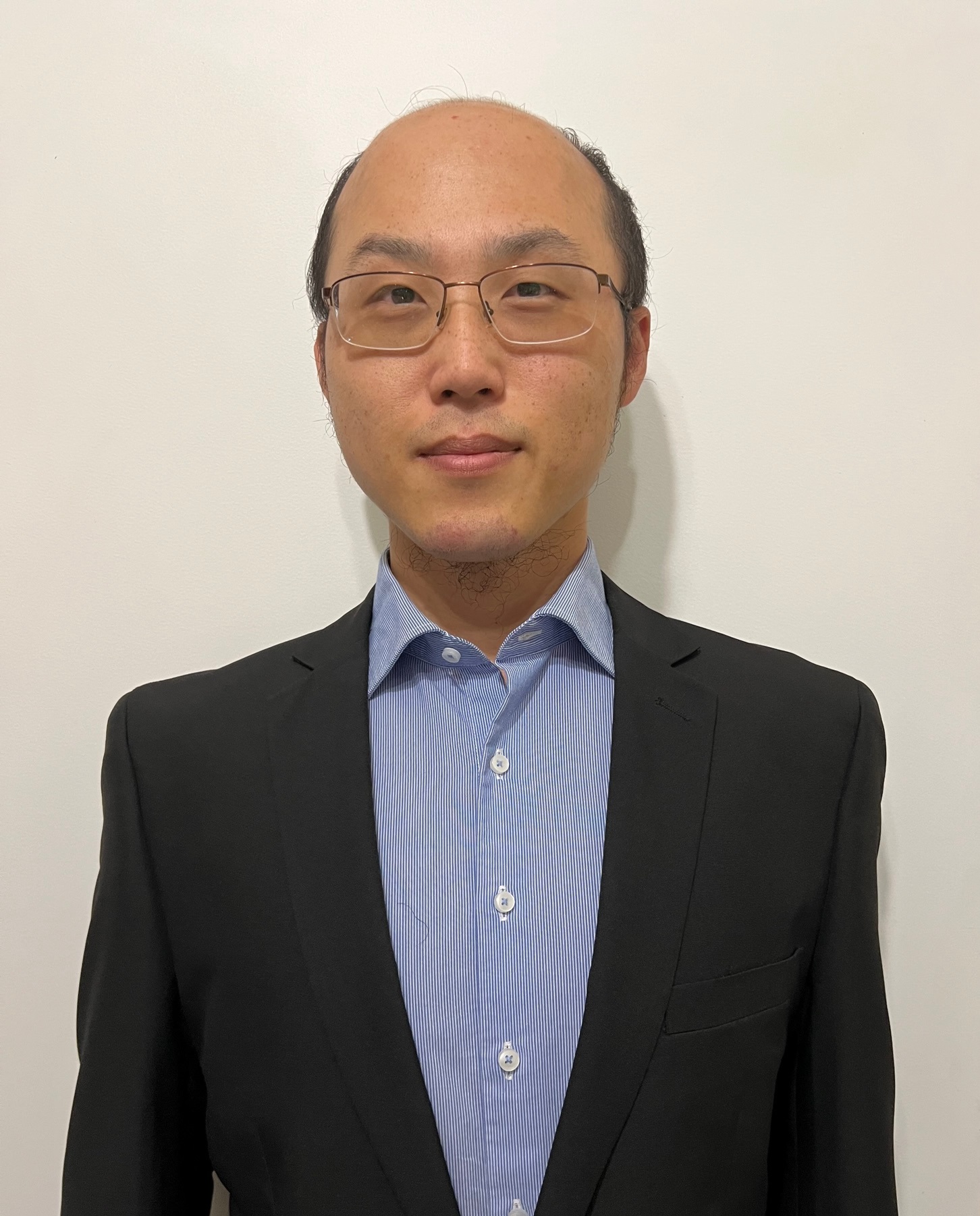}}]{Ruoxi Sun} received the Ph.D. degree with The University of Adelaide, Adelaide, SA, Australia. He is a Research Fellow with CSIRO’s Data61, Marsfield, NSW, Australia. His work in the cybersecurity and privacy domains has led to the publication of over ten papers in leading conferences and journals, such as the IEEE Symposium on Security and Privacy (S\&P), ACM Conference on Computer and Communications Security (CCS), Network and Distributed System Security Symposium (NDSS), The Web Conference (WWW), International Conference on Software Engineering (ICSE), The ACM Joint European Software Engineering Conference and Symposium on the Foundations of Software Engineering (ESEC/FSE), The IEEE/ACM Automated Software Engineering Conference (ASE), Annual Conference on Neural Information Processing Systems (NeurIPS), and others. His research interests include mobile security and privacy, the Internet of Things (IoT) security, and machine learning security.
\end{IEEEbiography}

\begin{IEEEbiography}
[{\includegraphics[width=1in,height=1.25in,keepaspectratio]{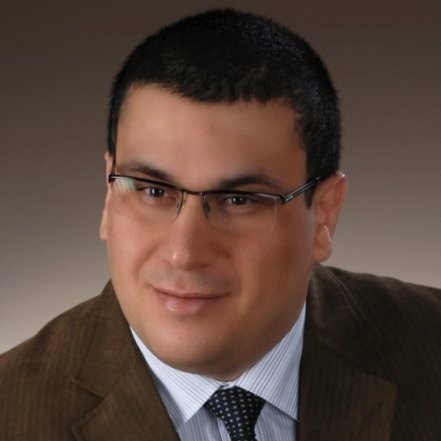}}]{Seyit Camtepe}
Dr Seyit Camtepe received his PhD in Computer Science from Rensselaer Polytechnic Institute (America's oldest technological research university), New York, USA 2007. From 2007 to 2013, he was a Senior Researcher and Team Leader focusing on AI and Cyber Security at the TU-Berlin (the first German university to adopt the name "Technische Universität"), Germany. He worked as an ECARD lecturer at the QUT, Australia, for five years. As part of these roles, he led or participated in many large impactful projects (i.e., multi-national and with more than five partners/countries) in Europe funded by the European Commission and in Australia, supported by industry and government. He completed CSIRO's Experienced Leader Program (ELP) and Spark Leader Labs to mature his leadership skills on top of the leadership skills he built over 27 years on three continents. He has been a PC member in prestigious ML and Cyber security conferences such as ESORICS, ASIACCS, AAAI and ECML-PKDD. He was a co-editor of a special issue on Cryptography: A Cyber Security Toolkit.
\end{IEEEbiography}

\begin{IEEEbiography}
[{\includegraphics[width=1in,height=1.25in,keepaspectratio]{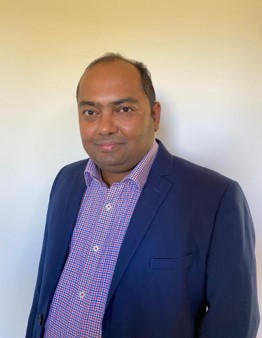}}]{Praveen Gauravaram}
Praveen has a PhD in Cryptology from Queensland University of Technology, Brisbane, Australia and has held scientific positions in India, Europe, and Australia and is a recipient of research grants and awards whist his research fellowship at Technical University of Denmark. Praveen was a recipient of young elite scientist award in 2010 from the Danish Agency for Science, Technology, and Innovation for his contributions to research in Cryptographic Hash Functions. Praveen has more than sixty publications and consulting advisories in cyber security. Praveen is professionally passioned about engaging customers in emerging technologies for customer business growth and successful digital transformation. Praveen holds honorary academic titles of Adjunct Associate Professor with the School of Computer Science and Engineering, Faculty of Engineering at UNSW and Adjunct Professor with the School of Information Technology, Faculty of Science, Engineering and Built Environment at Deakin University. Praveen is also an Advisor for the ICT Curriculum Advisory Group of Southern Cross University. Praveen enjoys playing weekend cricket, daily morning walks, and Aussie outdoors.
\end{IEEEbiography}

\begin{IEEEbiography}
[{\includegraphics[width=1in,height=1.25in,keepaspectratio]{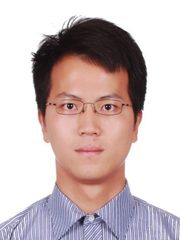}}]{Huaming Chen} (Member, IEEE) received the Ph.D. degree from the University of Wollongong, Wollongong, Australia. He is currently a senior lecturer with the School of Electrical and Computer Engineering, the University of Sydney, Sydney, Australia. His main research interests include software engineering/security, trustworthy AI, and applied machine learning. He is the organiser of TRUSTWORTHY AND RESPONSIBLE AI workshop. He serves on the program committees of ACM MM, CCS, IJCAI, KDD, The Web Conference, SIAM ICDM, ECML/PKDD and so on. 
\end{IEEEbiography}
\end{document}